\documentclass[prodmode,acmtecs]{acmsmall} % Aptara syntax

\usepackage{graphicx}
\usepackage{amssymb,amsfonts,amsmath}
\usepackage{float}
\usepackage{xspace}
\usepackage{times}
\usepackage{hyperref}
\usepackage{color}

\usepackage{graphicx}

\usepackage{algorithm}
\usepackage[noend]{algpseudocode}
\usepackage{IEEEtrantools}
\usepackage{xspace}

\usepackage{graphicx}
\usepackage{graphicx}
\usepackage{caption}
\usepackage{multirow}
\usepackage{epstopdf}
\usepackage{float}
\usepackage{amsmath, amssymb}
\usepackage[T1]{fontenc}
\usepackage{array}
\usepackage{makecell}
\usepackage{rotating}
\usepackage{mathtools}
\usepackage[table,xcdraw]{xcolor}
\usepackage{array,booktabs}
\usepackage{tikz}
\usepackage{xspace}

\usepackage{times}

\usetikzlibrary{matrix}
\usetikzlibrary{decorations.pathmorphing}
\usetikzlibrary{arrows,calc}

\usepackage{subcaption}

\newcommand{\mixed}{\textit{BuSca}\xspace}
\newcommand{\mixedext}{\textit{Burstiness Scale}\xspace}
\newcommand{\paramScale}{\psi}

\begin{document}

% Title portion
\title{Burstiness Scale: a highly parsimonious model for characterizing random series of events}

\author{
Rodrigo A S Alves
\affil{CEFET-MG}
Renato Assun\c{c}\~{a}o 
\affil{UFMG}
Pedro O.S. Vaz de Melo
\affil{UFMG}
}

\begin{abstract}
The problem to accurately and parsimoniously characterize random series of events (RSEs) present in the Web, such as e-mail conversations or Twitter hashtags, is not trivial. Reports found in the literature reveal two apparent conflicting visions of how RSEs should be modeled. From one side, the Poissonian processes, of which consecutive events follow each other at a relatively regular time and should not be correlated. On the other side, the self-exciting processes, which are able to generate bursts of correlated events and periods of inactivities. The existence of many and sometimes conflicting approaches to model RSEs is a consequence of the unpredictability of the aggregated dynamics of our individual and routine activities, which sometimes show simple patterns, but sometimes results in irregular rising and falling trends. In this paper we propose a highly parsimonious way to characterize general RSEs, namely the \mixedext (\mixed) model. \mixed views each RSE as a mix of two independent process: a \textit{Poissonian} and a self-exciting one. Here we describe a fast method to extract the two parameters of \mixed that, together, gives the burstyness scale $\paramScale$, which represents how much of the RSE is due to bursty and viral effects. We validated our method in eight diverse and large datasets containing real random series of events seen in Twitter, Yelp, e-mail conversations, Digg, and online forums. Results showed that, even using only two parameters, \mixed is able to accurately describe RSEs seen in these diverse systems, what can leverage many applications.
\end{abstract}

%\begin{bottomstuff}
%This work is partially supported by the author's individual grants and scholarships from CNPq and FAPEMIG.
%\end{bottomstuff}

\maketitle

\section{Introduction}
\label{sec:intro}

What is the best way to characterize random series of events (RSEs) present in the Web, such as Yelp reviews or Twitter hashtags?  Descriptively, one can characterize a given RSE as constant and predictable for a period, then bursty for another, back to being constant and, after a long period, bursty again. Formally, the answer to this question is not trivial. 
It certainly must include the extreme case of the homogeneous \textit{Poisson Process} (PP)~\cite{haight:1967}, which 
has a single and intuitive rate parameter $\lambda$. Consecutive events of PP follow each other at a relatively regular time and $\lambda$ represents the constant rate at which events arrive. The class of \textit{Poissonian} or \textit{completely random} processes includes also the case when $\lambda$ varies with time. In this class, events
must be without any aftereffects, that is, there is no interaction between any sequence of events~\cite{Snyder1991}. There are RSEs seen in the Web that were accurately modeled by a Poissonian process, such as many instances of viewing activity on Youtube~\cite{Crane2008}, e-mail conversations~\cite{malmgren:2008} and hashtag posts on Twitter~\cite{Lehmann2012}.

Unfortunately, recent analyzes showed that this simple and elegant model has proved unsuitable for many cases~\cite{oliveira:2005,eckmann:2004,VazdeMelo2015,vazdemelo:2013a}. Such analyzes revealed that many RSEs produced by humans have very long periods of inactivity and also bursts of intense activity~\cite{barabasi:2005,Jiang:2005}, in contrast to Poissonian processes, where activities may occur at a fairly constant rate. Moreover, many RSEs in the Web also have strong correlations between historical data and future data~\cite{Crane2008,VazdeMelo2015,vazdemelo:2013a,FerrazCosta2015,Masuda2013}, a feature that must not occur in Poissonian processes. These RSEs fall into a particular class of random point processes, the so called \textit{self-exciting} processes~\cite{Snyder1991}. The problem of characterizing such RSEs is that they occur in many shapes and in very unpredictable ways~\cite{vazdemelo:2013a,Matsubara2012,Figueiredo2014,FerrazCosta2015,Yang2011,Crane2008}. They have the so called ``quick rise-and-fall'' property~\cite{Matsubara2012} of bursts in cascades, producing correlations between past and future data that becomes difficult to be captured by regression-based methods~\cite{AAAI159338}. 

%So, in the one side of the spectrum we have a class of well behaved RSEs that are easily and parsimoniously modeled by Poissonian processes. On the other side, we have self-excited and noisy RSEs, which are difficult to predict and to characterize accurately in a parsimonious fashion. In this context, we ask: should all random series of events must be in either side of this spectrum? If not, how can we identify and parsimoniously characterize those that fall in the middle? Is there a simple way to position a given RSE in this spectrum? Finally, what are the benefits and applications that one can leverage from this approach?

\begin{figure*}[!htb]
 \centering
        \begin{subfigure}[b]{0.30\textwidth}
  	  \includegraphics[width=\textwidth]{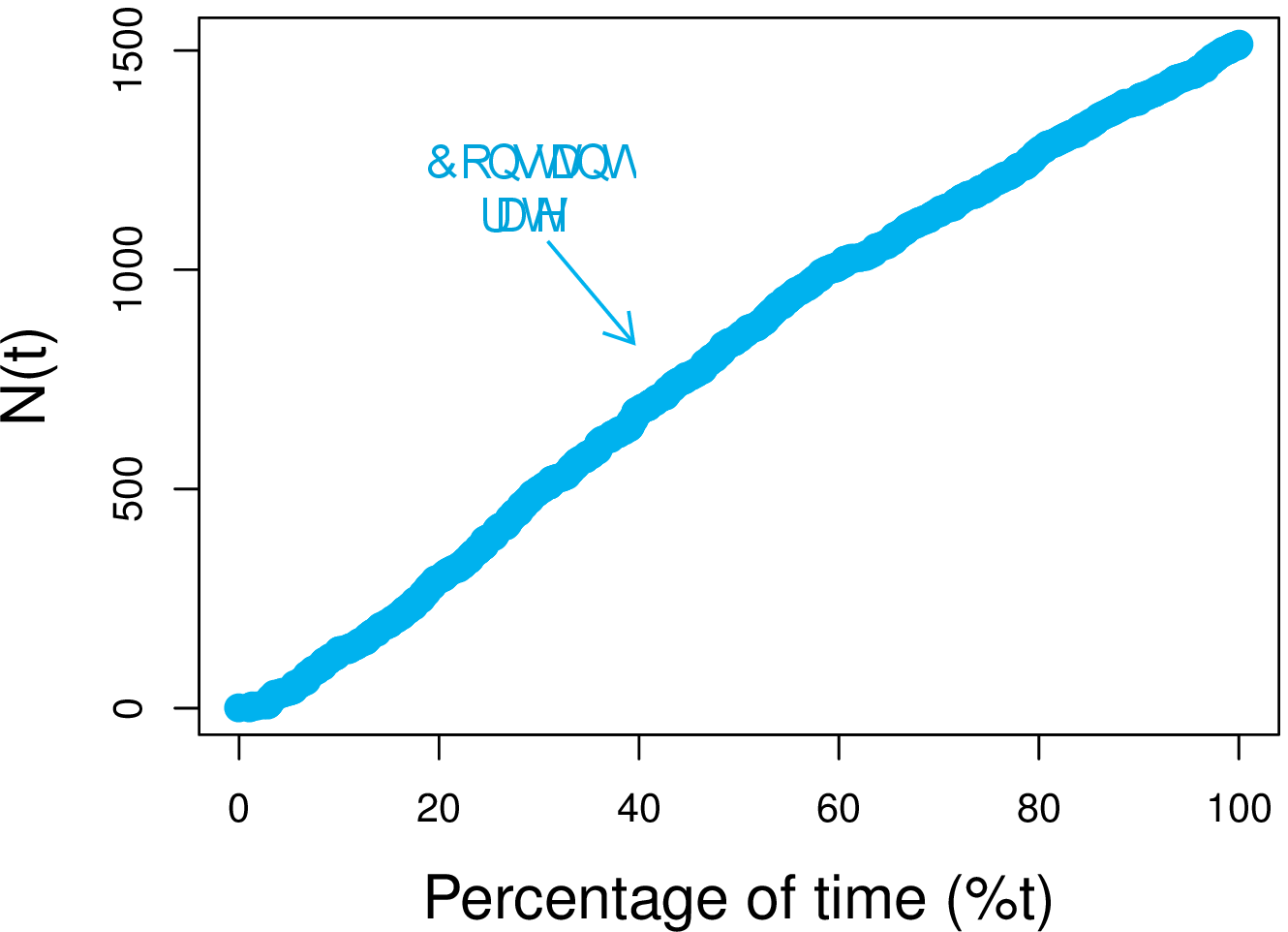}
                \caption{\#wheretheydothatat}
                \label{fig:poisson_introduction}
        \end{subfigure}%
        ~ %add desired spacing between images, e. g. ~, \quad, \qquad, \hfill etc.
        \begin{subfigure}[b]{0.30\textwidth}
               \includegraphics[width=\textwidth]{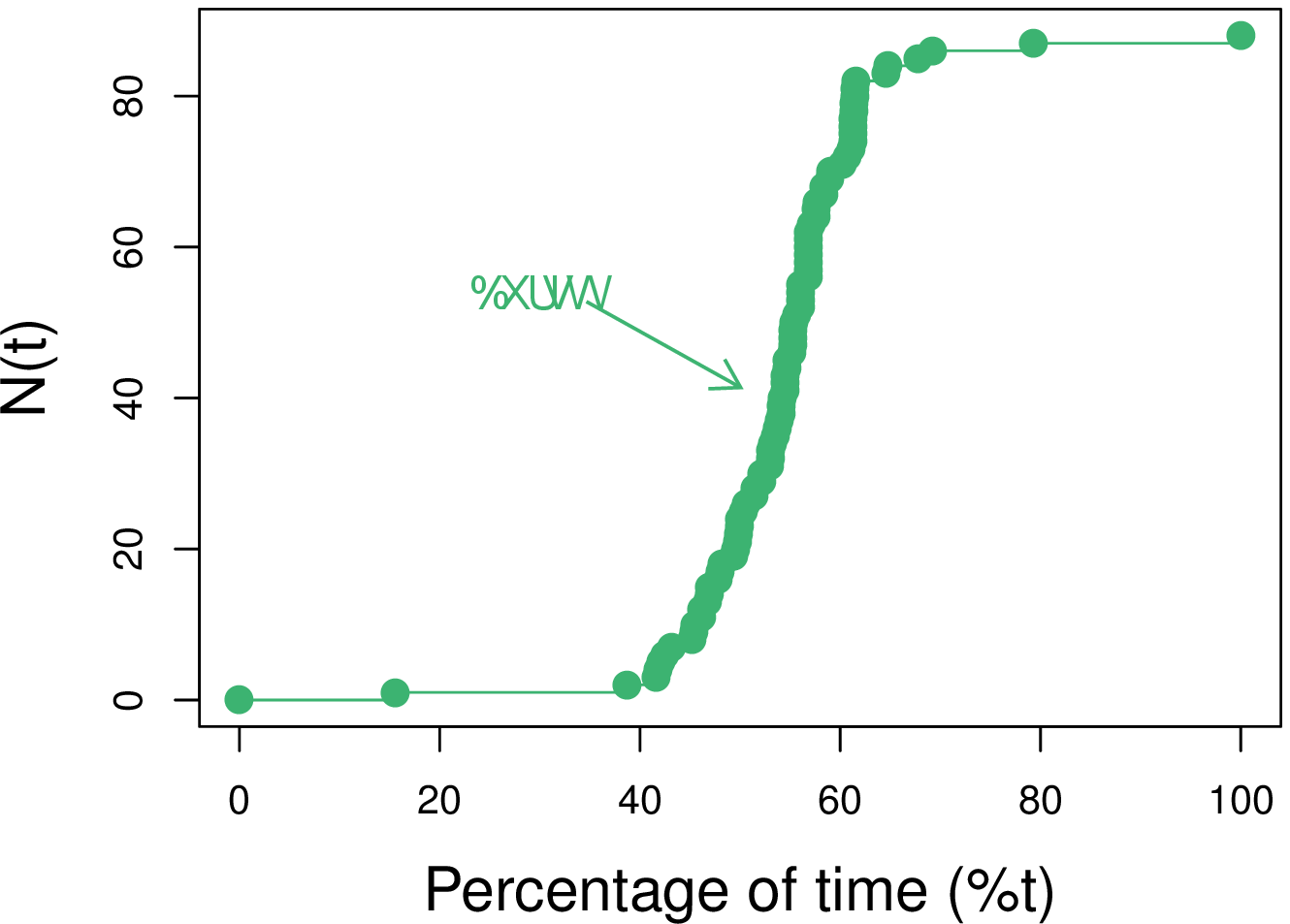}
                \caption{\#cotto}
                \label{fig:sfp_introduction}
        \end{subfigure}
        ~ %add desired spacing between images, e. g. ~, \quad, \qquad, \hfill etc.
        \begin{subfigure}[b]{0.30\textwidth}
                 \includegraphics[width=\textwidth]{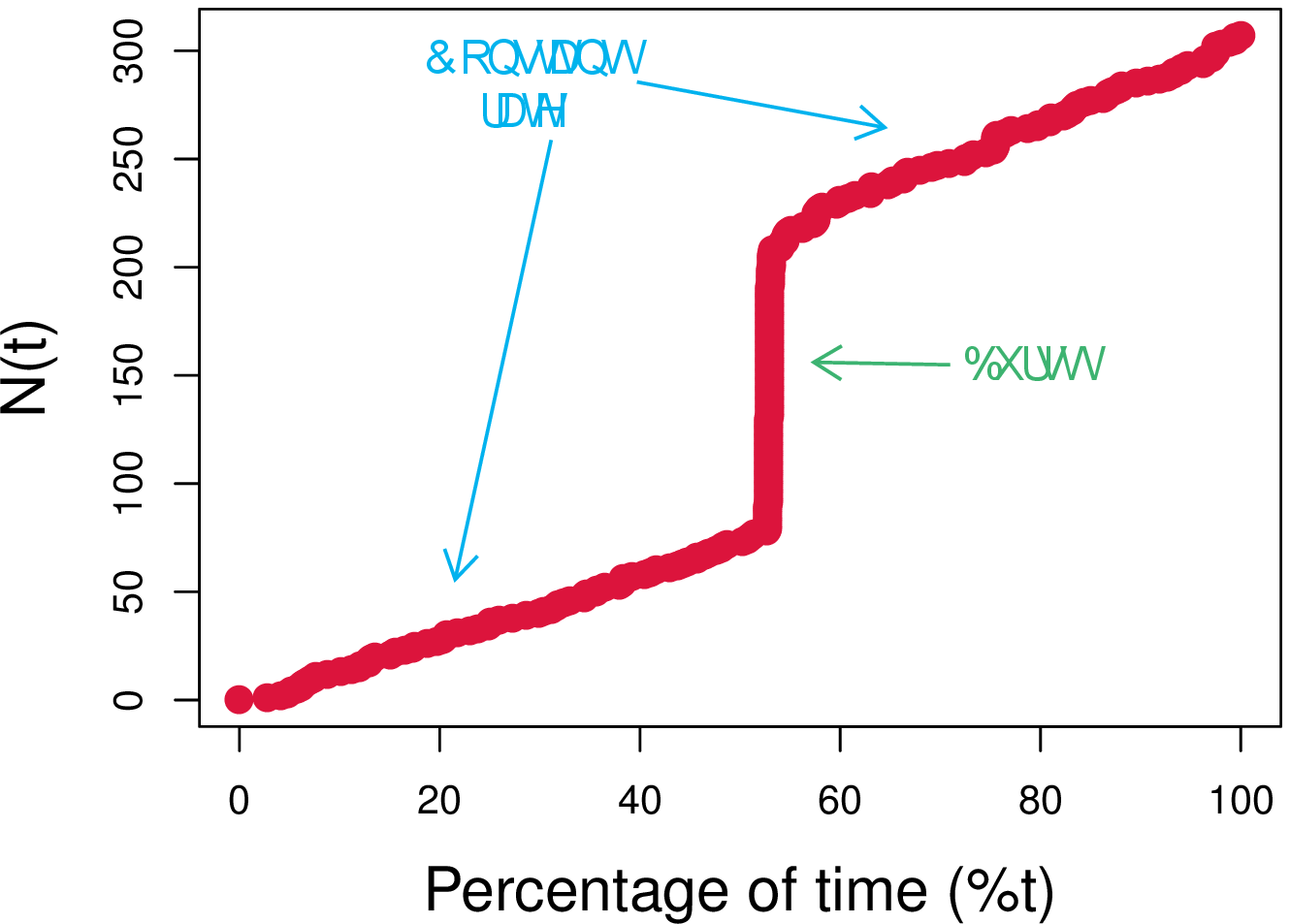}
                \caption{\#ta}
                \label{fig:mixture_introduction}
        \end{subfigure}
    \caption{Three reals individuals of twitter database  }\label{fig:introduction_pics}
 
\end{figure*}

As pointed out by~\cite{Crane2008}, the aggregated dynamics of our individual activities is a consequence of a myriad of factors that guide individual actions, which produce a plethora of collective behaviors. Thus, in order to accurately capture all patterns seen in human-generated random series of events, researchers are proposing models with many parameters and, for most of the times, tailored to a specific activity in a specific  system~\cite{malmgren:2009,Yang2011,Lehmann2012,ICWSM1510537,AAAI159338,Matsubara2012,Figueiredo2014,FerrazCosta2015,Masuda2013,Pinto2015,Zhao2015,yang13a}. Going against this trend, in this work we propose the \mixedext (\mixed) model, a highly parsimonious model to characterize RSEs that can be (i) purely Poissonian, (ii) purely self-exciting or (iii) a mix of these two behaviors. In \mixed, the underlying Poissonian process is responsible for the arrival of events related to the routine activity dynamics of individuals, whereas the underlying self-exciting process is responsible for the arrival of bursty and ephemeral events, related to the endogenous (e.g. online social networks) and the exogenous (e.g. mass media) mechanisms that drive public attention and generate the ``quick rise-and-fall'' property and correlations seen in many RSEs~\cite{Crane2008,Lehmann2012}. To illustrate that, observe  Figure~\ref{fig:introduction_pics}, which shows the cumulative number of occurrences $N(t)$ of three Twitter hashtags over time. In Figure~\ref{fig:introduction_pics}a, the curve of hashtag \texttt{\#wheretheydothatat} is a straight line, indicating that this RSE is well modeled by a PP. In Figure~\ref{fig:introduction_pics}b, the curve of hashtag \texttt{\#cotto} has long periods of inactivities and a burst of events, suggesting that the underlying process may be self-exciting in this case. 
%Moreover, as seen in the inset, the odds ratio is a power law, indicating that this RSE is well modeled by a pure SFP~\cite{vazdemelo:2013a}. 
Finally, In Figure~\ref{fig:introduction_pics}c, the curve of hashtag \texttt{\#ta} is apparently a mix of these two processes, exactly what \mixed aims to model. 
 
%In order to answer to these questions, we propose the \mixedext (\mixed) model, which views each random series of events as a mix of two independent process: a \textit{Poissonian} and a self-exciting one. As pointed out by~\cite{Crane2008}, the aggregated dynamics of our individual and routine activities create seasonal trends or simple patterns, but sometimes our collective action results in blockbusters, best sellers, and other large-scale trends in financial and cultural markets. Here we argue that these two dynamics do not need to occur separately, but may coexist in general random series of events, particularly in the Web. In \mixed, the underlying Poissonian process is responsible for the arrival of events related to the routine and seasonal activity dynamics of individuals, whereas the underlying self-exciting process is responsible for the arrival of bursty and trendy events, related to the endogenous (e.g. online social networks) and the exogenous (e.g. mass media) mechanisms that drive public attention and generate the ``quick rise-and-fall'' property and correlations seen in many RSEs~\cite{Crane2008,Lehmann2012}.

Besides that, our goal is also to characterize general RSEs using the least amount of parameters possible. The idea is to propose a highly parsimonious model that can separate out constant and routine events from bursty and ephemeral events in general RSEs. We present and validate a particular and highly parsimonious case of \mixed, where the Poissonian process is given by a homogeneous Poisson process and the self-exciting process is given by a Self-Feeding Process (SFP)~\cite{vazdemelo:2013a}. We chose these models because (i) both of them require a single parameter and (ii) they are on opposite ends of the spectrum. The PP is on the extreme side where the events do not interact with each other and inter-event times are independent. On the other extreme lies the SFP, where consecutive inter-event times are highly correlated. Even though \mixed has only two parameters, we show that, surprisingly, it is is able to accurately characterize a large corpus of diverse RSE seen in Web systems, namely Twitter, Yelp, e-mails, Digg, and online forums. We show that disentangling constant from ephemeral events in general RSEs may reveal interesting, relevant and fascinating properties about the underlying dynamics of the system in question in a very summarized way, leveraging applications such as monitoring systems, anomaly detection methods, flow predictors, among others. 

%To illustrate that, observe  Figure~\ref{fig:introduction_pics}, which shows the cumulative number of occurrences of Twitter hashtags over time. In Figure~\ref{fig:introduction_pics}a, the curve of hashtag \texttt{#wheretheydothatat} is a straight line, indicating that this RSE is well modeled by a PP. In Figure~\ref{fig:introduction_pics}b, the curve of hashtag \texttt{#cotto} has long periods of inactivities and a burst of events. Moreover, as seen in the inset, the odds ratio is a power law, indicating that this RSE is well modeled by a pure SFP~\cite{vazdemelo:2013a}. Finally, In Figure~\ref{fig:introduction_pics}c, the curve of hashtag \texttt{#ta} is apparently a mix of these two processes, suggesting that \mixed is an appropriate model for it. 

%Besides Twitter data, we validate our approach by analyzing other seven diverse and large datasets containing real random series of events seen in, for instance, Yelp, Digg, and online forums. Disentangling a Poissonian and a self-exciting processes from a mixed one in this plethora of RSEs is a challenging task but, nevertheless, may reveal interesting, relevant and fascinating properties about the underlying dynamics of the system in question. Moreover, \mixed offers a very summarized and intuitive way to characterize a RSE, which and, consequently, may leverage several applications, such as monitoring systems, anomaly detection methods, flow predictors, among others.

In summary, the main contributions of this paper are:

\begin{itemize}
	\item \mixed, a widely applicable model that parsimoniously characterize communication time series with only two intuitive parameters and validated in eight different datasets. From these parameters, we can calculate the \textit{burstiness scale} $\paramScale$, which represents how much of the process is due to bursty and viral effects.
	\item A fast and scalable method to separate events arising from a homogeneous Poisson Process from those arising from a self-exciting process in RSEs.
	\item A method to detect anomalies and another to detect bursts in random series of events.
\end{itemize}

The rest of the paper is organized as follows. In Section~\ref{sec:related}, 
we provide a brief survey of the related work. Our model is introduced in Section~\ref{sec:ourmodel} 
together with the algorithm to estimate its parameters. We show that the maximum likelihood estimator 
is biased and show to fix the problem in Section~\ref{sec:mlebias}, discussing a statistical test   
procedure to discriminate between extreme cases in Section~\ref{sec:classtest}.
In Section~\ref{sec:dados}, we describe the eight datasets used in this work and show the goodness of fit 
of our model in Section~\ref{sec:goodness}. A comparison with the Hawkes model is given in 
Section~\ref{sec:Hawkes}. We show two applications of our model in Section~\ref{sec:aplications}.
We close the paper with Section~\ref{sec:conclusions}, where we  present our conclusions.

\section{Related work}
\label{sec:related}

%interest in understanding the dynamics of random series of events in the Web. Some models. 
Characterizing the dynamics of human activity in the Web has attracted the attention of the research
community~\cite{barabasi:2005,Crane2008,malmgren:2009,Wang2012,Yang2011,Lehmann2012,ICWSM1510537,AAAI159338,Matsubara2012,Figueiredo2014,FerrazCosta2015,Masuda2013,Pinto2015,Zhao2015,yang13a} as it has implications that can benefit a large number of applications, such as trend detection~\cite{Pinto2015}, popularity prediction~\cite{Lerman2010}, clustering~\cite{Du2015}, anomaly detection~\cite{VazdeMelo2015}, among others. The problem is that uncovering the rules that govern human behavior is a difficult task, since many factors may  influence an individual's decision to take action. Analysis of real data have shown that human activity in the Web can be highly unpredictable, ranging from being completely random~\cite{Crane2008,Cao:2001,Karagiannis:2004,kleinberg:2002,malmgren:2008,malmgren:2009} to highly correlated and bursty~\cite{barabasi:2005,Jiang:2005,vazdemelo:2013a,Matsubara2012,Masuda2013,yang13a,Pinto2015,Zhao2015}.

%since analysis of real data have shown that humans have very long periods of inactivity and bursts of intense activity~\cite{barabasi:2005, Jiang:2005}, in contrast to the PP, where activities occur at a fairly constant rate. Although researches agree that the PP is not suitable, there is no consensus about the right model between two major schools of thought. The first viewpoint~\cite{barabasi:2005,oliveira:2005} states that a power law~\cite{faloutsos:1999} is an appropriate fit for the Probability Density Function (PDF) of the \textit{inter-event time distribution} (\ied{}), where bursts and heavy-tails in human activities are a consequence of a decision-based queuing process, when tasks are executed according to some perceived priority. The second viewpoint is that the \ied is well explained by variations of the PP~\cite{kuczura:1973,malmgren:2008,malmgren:2009,malmgren:2009b,kleinberg:2002}. They are based on the fact that short-term communication events exhibits a Poissonian behavior~\cite{Cao:2001,Karagiannis:2004} and suggest a piece-wise Poisson process: the first interval has a constant rate $\lambda$; for the next, change the rate, and continue. 

As one of the first attempts to model bursty RSE, Barab\'{a}si et. al.~\cite{barabasi:2005} proposed that bursts and heavy-tails in human activities are a consequence of a decision-based queuing process, when tasks are executed according to some perceived priority. In this way, most of the tasks would be executed rapidly while some of them may take a very long time. The queuing models generate power law distributions, but do not correlate the timing of events explicitly. As an alternative to queuing models, many researchers started to consider the self-exciting point processes, which are also able to model correlations between historical and future events. In a pioneer effort, Crane and Sornette~\cite{Crane2008} modeled the viewing activity on Youtube as a Hawkes process. They proposed that the burstiness seen in data is a response to endogenous word-of-mouth effects or sudden exogenous perturbations. This seminal paper inspired many other efforts to model human dynamics in the Web as a Hawkes process~\cite{Matsubara2012,Masuda2013,yang13a,Pinto2015,Zhao2015}. Similar to the Hawkes process, the Self-Feeding process (SFP)~\cite{vazdemelo:2013a} is another type of self-exciting process that also captures correlations between historical and future data, being also used to model human dynamics in the Web~\cite{vazdemelo:2013a,VazdeMelo2015,FerrazCosta2015}. Different from Hawkes, whose conditional intensity explicitly depends on all previous events, the SFP considers only the previous inter-event time.

%poissonian evidences
Although there are strong evidences that self-exciting processes are well suited to model human dynamics in the Web, there are studies that show that the Poisson process and its variations are also appropriate~\cite{Crane2008,Cao:2001,Karagiannis:2004,kleinberg:2002,malmgren:2008,malmgren:2009}. \cite{Cao:2001,Karagiannis:2004} showed that Internet traffic can be accurately modeled by a Poisson process under particular circustances, e.g. heavy traffic. When analyzing Youtube viewing activity,~\cite{Crane2008} verified that 90\% of the videos analyzed either do not experience much activity or can be described statistically as a Poisson process. Malmgreen et\ {al.}~\cite{malmgren:2008,malmgren:2009} showed that a non-homogeneous Poisson process can accurately describe e-mail communications. In this case, the rate $\lambda(t)$ varies with time, in a periodic fashion (e.g., people answer e-mails in the morning; then go to lunch; then answer more e-mails, etc).  

%reconciling all theories: our model
These apparently conflicting approaches, i.e., self-exciting and Poissonian approaches, motivated many researchers to investigate and characterize this plethora of human behaviors found in the Web. For instance, ~\cite{Vallet:2015:CPV:2806416.2806556} used a machine learning approach to characterize videos on Youtube. From several features extracted from Youtube and Twitter, the authors verified that the current tweeting rate along with the volume of tweets since the video was uploaded are the two most important Twitter features for classifying a Youtube video into viral or popular. In this direction, \cite{Lehmann2012} verified that Twitter hashtag activities may by continuous, periodic or concentrated around an isolated peak, while \cite{Figueiredo2014} found that revisits account from 40\% to 96\% of the popularity of an object in Youtube, Twitter and LastFm, depending on the application. ~\cite{ICWSM1510537} verified that the popularity of a Youtube video can go through multiple phases of rise and fall, probably generated by a number of different background random processes that are super-imposed onto the power-law behavior.
%They also observed that, surprisingly, older videos are as likely to receive as much attention as new ones.
The main difference between these models and ours is that the former ones mainly focus on representing all the details and random aspects of very distinct RSEs, which naturally demands many parameters. In our case, our proposal aims to disentangle the bursty and constant behavior of RSEs as parsimoniously as possible. Surprisingly, our model is able to accurately describe a large and diverse corpus of RSEs seen in the Web with only two parameters.

In this version, random series of events are modeled by a mixture of two independent processes: a Poisson process, which accounts for the background constant behavior, and a one-parameter SFP, which accounts for the bursty behavior. A natural question that arises is: how different is this model from the widely used Hawkes process? The main difference are twofold. First, in the Hawkes process, every single arriving event excites the process, i.e., is correlated to the appearance of future events. In our proposal, since the PP is independent, non-correlated events may arrive at any time. Second, our model is even more parsimonious than the Hawkes process, two parameters against three\footnote{Considering the most parsimonious version of the Hawkes process.}. In Section~\ref{sec:Hawkes} we quantitatively show that our proposed model is more suited to real data than the Hawkes process.

\section{Modeling information bursts}
\label{sec:ourmodel}

Point processes is the stochastic process framework developed to model a random sequence of events (RSE). 
Let $0 <  t_1 < t_2 < \ldots$ be a sequence of random event times, with $t_i \in \mathbb{R}^+$, and 
$N(a,b)$ be the random number of events in $(a,b]$. We simplify the notation when the interval starts on 
$t=0$ by writing simply $N(0, b) = N(b)$. Let $\mathcal{H}_t$ be the random history of the process up to, 
but not including,
time $t$. A fundamental tool for modeling and for inference in point processes is the conditional intensity rate function.
It completely characterizes the distribution of the point process and it is given by
\begin{equation}
\lambda(t | \mathcal{H}_t) = \lim_{h\to 0} \frac{\mathbb{P}\left( N(t,t+h)  > 0 |  \mathcal{H}_t \right) }{h}  = 
\lim_{h\to 0} \frac{\mathbb{E}\left( N(t,t+h) |  \mathcal{H}_t \right) }{h}\: . 
\label{eq:cond_intens}
\end{equation}
The interpretation of this random function is that, for a small time interval $h$, the value of
$\lambda(t | \mathcal{H}_t) \times h$ is approximately 
the expected number of events in $(t, t+h)$. It can also be interpreted as the probability that the interval 
$(t, t+h)$ has at least one event given the random history of the process up to $t$.   
The notation emphasizes that the conditional intensity at time $t$ depends on the random events that occur previous to $t$. 
This implies that $\lambda(t | \mathcal{H}_t)h$ is a random function rather than a typical mathematical function.
The unconditional and non-random function $\lambda(t) = \mathbb{E}\left( \lambda(t | \mathcal{H}_t) \right) $
is called the intensity rate function where the expectation is taken over all possible process histories.  

The most well known point process is the Poisson process where $\lambda(t | \mathcal{H}_t) = \lambda(t)$. 
That is, the occurrence rate varies in time but it is deterministic such as, for example, 
$\lambda(t) = \beta_0 + \beta_1 \sin(\omega t)$. 
The main characteristics
of a Poisson process is that the counts in disjoint intervals are independent random variables with $N(a,b)$ having 
Poisson distribution with mean given by $\int_a^b \lambda(t) dt$. 
When the intensity does not vary in time, with $\lambda(t) \equiv \lambda$, we have a homogeneous 
Poisson process. 

\subsection{Self-feeding process}

The self-feeding process (SFP)~\cite{vazdemelo:2013a}  conditional intensity 
has a simple dependence on its past. 
Locally, it acts as a homogeneous 
Poisson process but its conditional intensity rate is inversely proportional to the temporal gap between the two last events. 
More specifically, the conditional intensity function is given by 
\begin{equation}
\lambda_s(t | \mathcal{H}_t) = \frac{1}{\mu/e + \Delta t_i} 
\label{eq:cond_intens_sfp}
\end{equation}
where $\Delta t_i = t_i - t_{i-1}$ and $t_i = \max_k \{ t_k : t_k \leq t \}$. This implies that the inter-event times 
$\Delta t_{i+1} = t_{i+1} - t_{i}$ are exponentially distributed with expected value $\mu/e + \Delta t_{i}$. 
The inter-event times $\Delta t_i$ follow a Markovian property. The constant $\mu$ is the median of the inter-event times 
and $e \approx 2.718$ is the Euler constant. A more general version of the SFP uses an additional parameter $\rho$ which was 
taken equal to 1 in this work. The motivation for this is that, in many databases analysed previously~\cite{vazdemelo:2013a} , it was found that $\rho \approx 1$.  An additional benefit of this decision is the simpler likelihood calculations involving the SFP. 

\begin{figure}[!htp]
\centering
        \begin{subfigure}[b]{.25\textwidth}
                 \includegraphics[width=\textwidth]{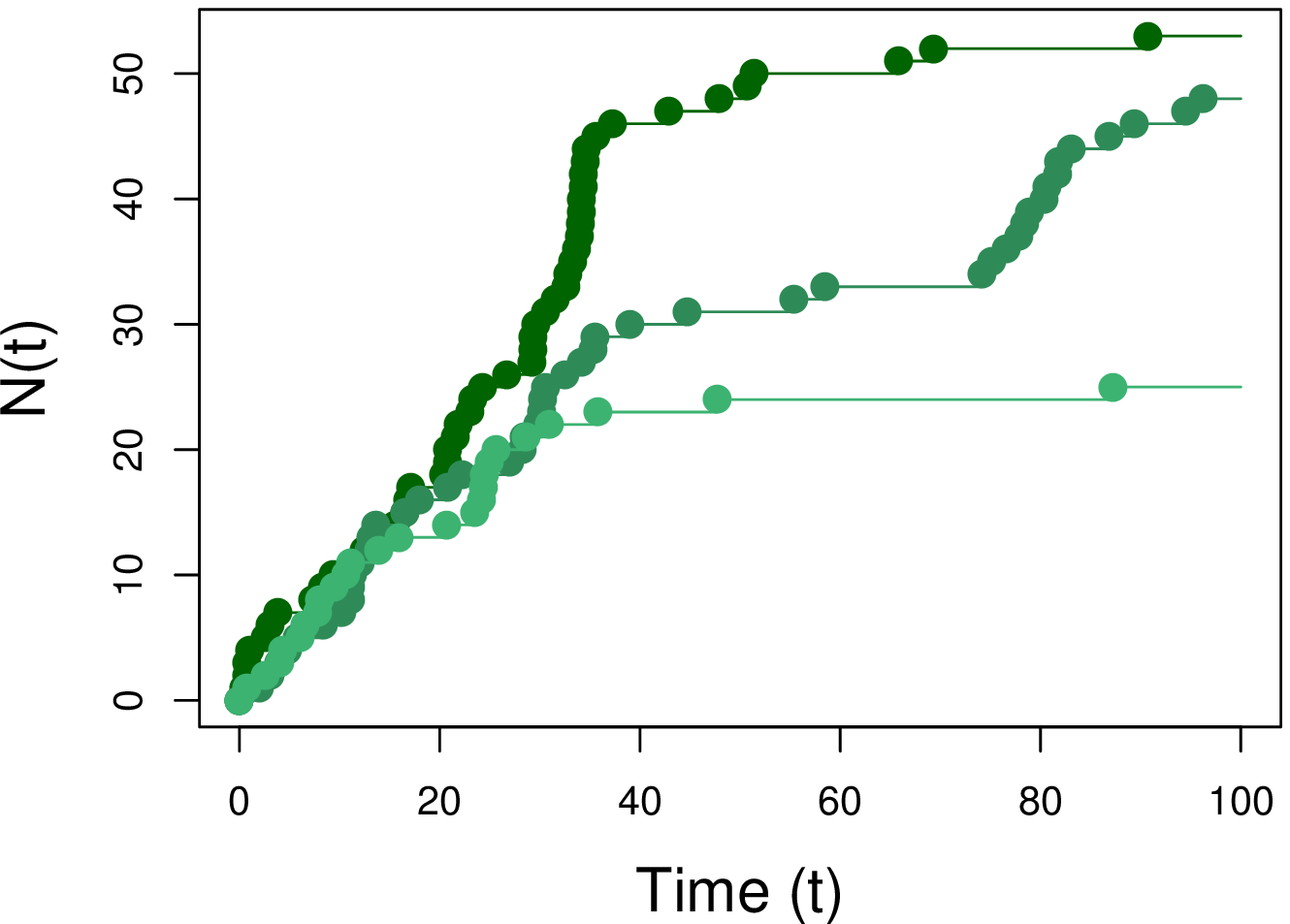}
                \caption{SFP realizations}
                \label{fig:sfp_final}
       \end{subfigure}%
%        ~ %add desired spacing between images, e. g. ~, \quad, \qquad, \hfill etc.
%           %(or a blank line to force the subfigure onto a new line)
        \begin{subfigure}[b]{0.25\textwidth}
                 \includegraphics[width=\textwidth]{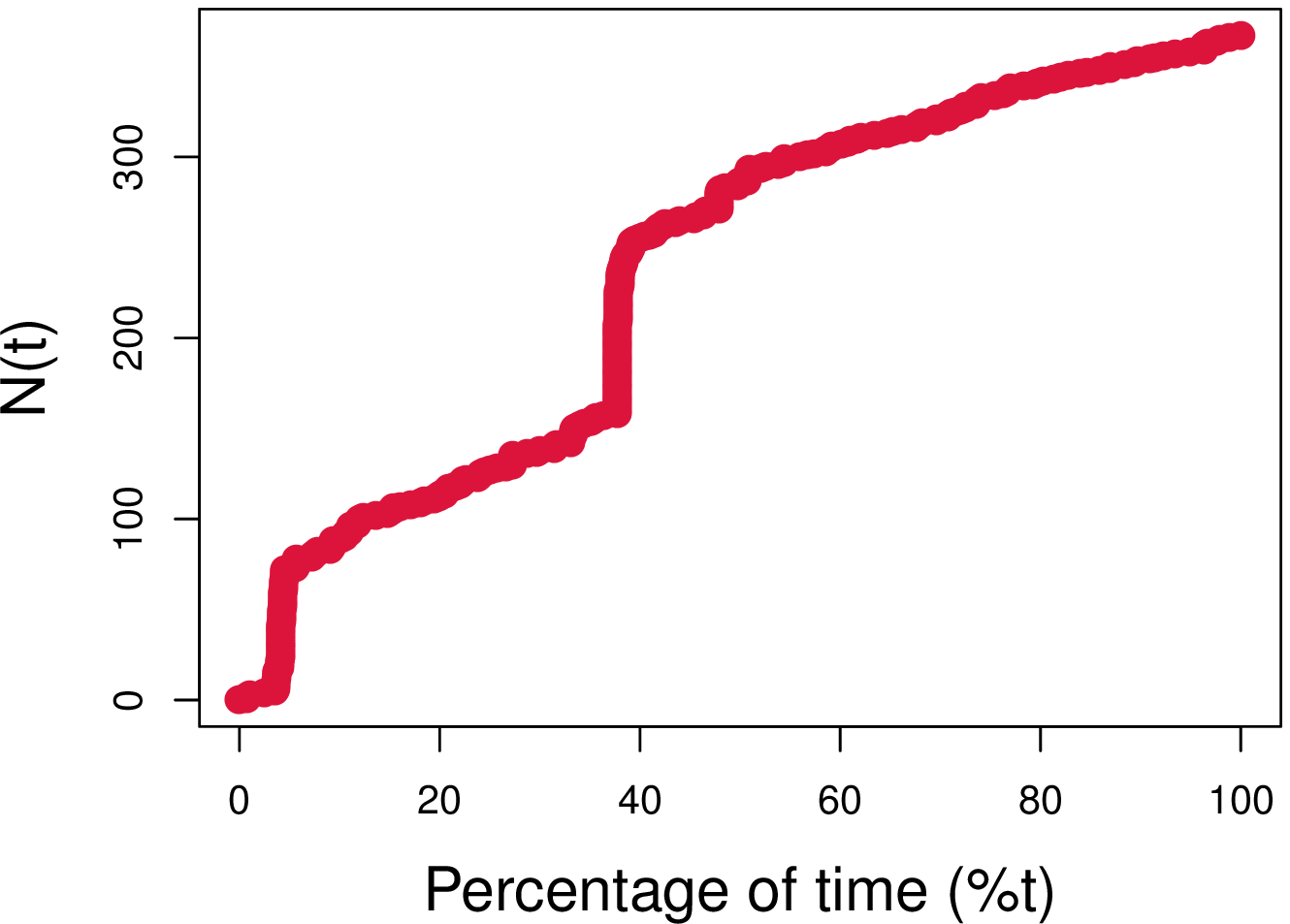}
                \caption{Twitter data}
                 \label{fig:mixture_2449_twitter}
         \end{subfigure}
    \caption{Three instances of the SFP process with $\mu = 1$ in the time interval $[0,100)$ (left) and a real time series with the \#shaq hashtag from Twitter.}
%         ~ %add desired spacing between images, e. g. ~, \quad, \qquad, \hfill etc.
          
\end{figure}

The Figure \ref{fig:sfp_final} presents three realizations of the SFP process in the interval $(0,100)$ with parameter $\mu = 1$. 
The vertical axis shows the accumulated number of events $N(t)$ up to time $t$. One striking aspect of this plot is its  variability.
In the first 40 time units, the lightest individual shows a rate of approximately 0.5 events per unit time while the darkest one has a rate of 2.25.
Having accumulated a very different number of events, they do not have many additional points after time $t=40$. The third one has a more 
constant rate of increase during the whole time period. Hence, with the same parameter $\mu$, we can see very different realizations 
from the SFP process. A common characteristic of the SFP instances is the mix of bursty periods alternating with quiet intervals.

% ?? Mostrar empiricamente que em algumas bases os silencios prolongados do SFP nao existem?? Tres individuos particulares.
% Mais tarde no paper, mostrar que o modelo de mistura ajusta-se bem neste tres individuos particulares??

However, in many datasets we have observed a departure from this SFP behavior. The most noticeable discrepancy is the absence of the 
long quiet periods predicted by the SFP model. To be concrete, consider the point processes realizations in Figure \ref{fig:mixture_2449_twitter}.
This plot is the cumulative counting of Twitter posts from the hashtag \texttt{shaq}. There are two clear bursts, when the series has a large 
increase in a short period of time. Apart from these two periods, the counts increase in a regular and almost constant rate. We do not 
observe long stretches of time with no events, as one would expect to see if a SFP process is generating these data. 

\subsection{The BuSca Model}

In this work, we propose a point process model that exhibits the same behavior consistently observed in our empirical findings: we want a 
mix of random bursts followed by more quiet periods, and we want realizations where the long silent periods predicted by the SFP 
are not allowed. To obtain these two aspects we propose a new model that is a mixture of the SFP process, to guarantee the presence of 
random bursts, with a homogeneous Poisson process, to generate a random but rather constant rate of events, breaking the long empty spaces 
created by the SFP. While the SFP captures the viral and ephemeral ``rise and fall'' patters, the PP captures the routine activities, acting as a random background noise added to a signal point process. We call this model the \mixedext (\mixed) model.

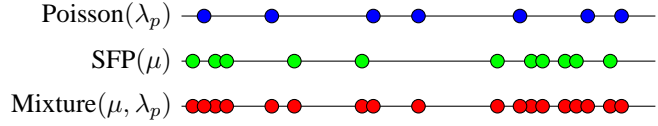
\begin{figure}[H]
\centering
\begin{tikzpicture}[scale=0.15]
    \begin{scope} % Energy levels
        \draw (1,0) node[left] {Mixture$(\mu,\lambda_{p})$} -- 
            ++(42,0);
        \draw (1,4) node[left] {SFP$(\mu)$} -- 
            ++(42,0); 
        \draw (1,8) node[left] {Poisson$(\lambda_{p})$} -- 
            ++(42,0);
     
    \end{scope}

    \begin{scope} % Electrons
        \foreach \x in {3,9,18,22,31,37,40}
            \filldraw[fill=blue] (\x, 8) circle (.6);
        \foreach \x in {2,4,5,11,17,29,32,33,35,36,39}
            \filldraw[fill=green] (\x, 4) circle (.6);
         \foreach \x in {2,4,5,11,17,29,32,33,35,36,39,3,9,18,22,31,37,40}
            \filldraw[fill=red] (\x, 0) circle (.6);
    \end{scope} 
    \end{tikzpicture}
	\caption{The \mixed model. The top line displays the events from the Poisson$(\lambda_p)$ component along the timeline while the middle line
	displays those from the SFP$(\mu)$ component. The user observes only the third line, the combination of the first two, without a label to identify the source process associated with each event.}\label{fig:mistura_diagrama}
\end{figure}

Figure \ref{fig:mistura_diagrama} shows the main idea of \mixed. The observed events are those on the bottom line. They are composed by two 
types of events, each one generated by a different point process. Each observed event can come either from a Poisson process (top line) or 
from an SFP process (middle line). We observe the mixture of these two types of events on the third line without an identifying label. This lack of knowledge of the
sourse process for each event is the cause of most inferential difficulties, as we discuss later. 

\begin{figure*}
 \centering
        \begin{subfigure}[b]{0.30\textwidth}
  	  \includegraphics[width=\textwidth]{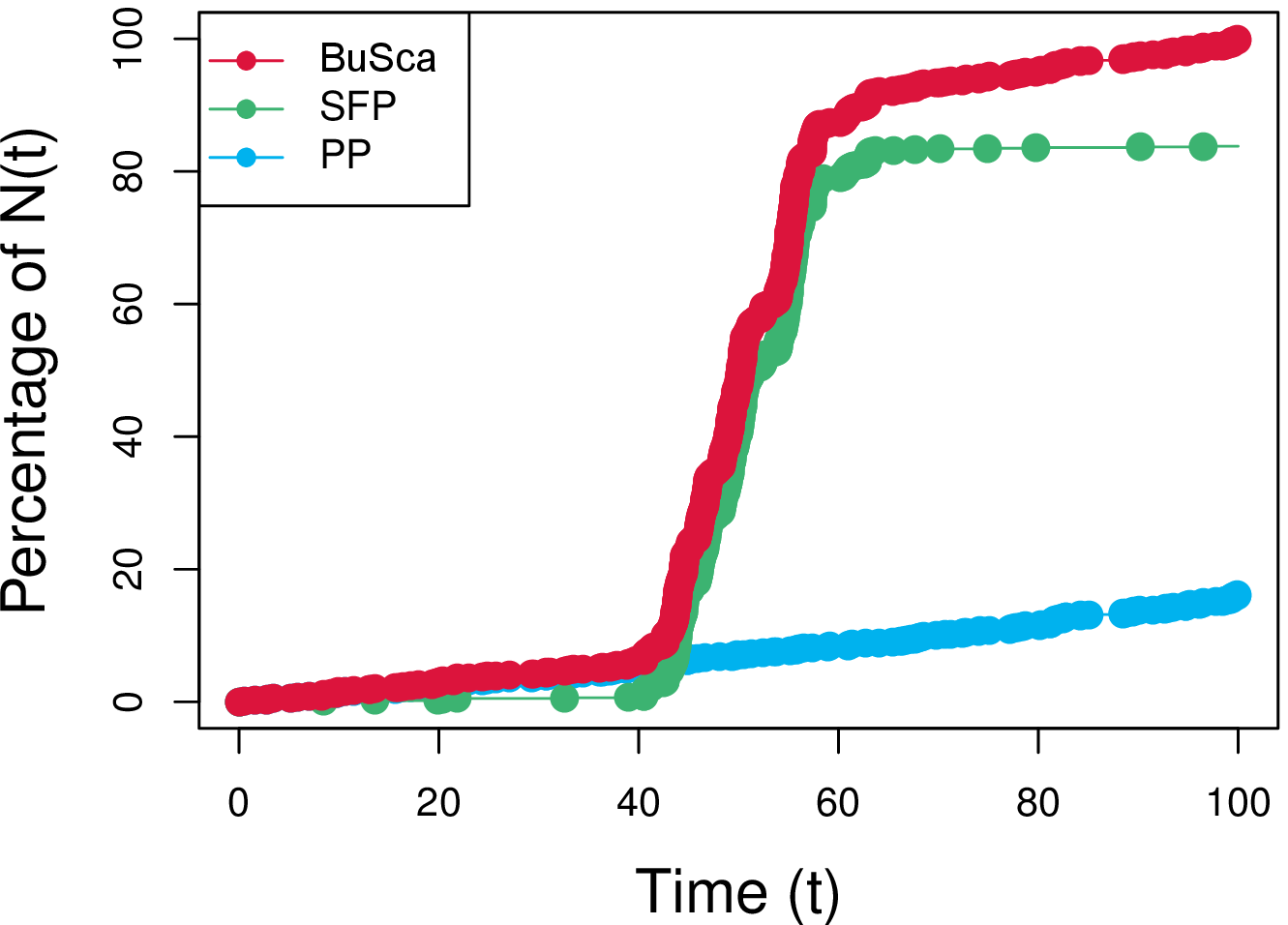}
                \caption{$\paramScale = 75$}
                \label{fig:mix_mu_100_lamb_010_1}
        \end{subfigure}%
        ~ %add desired spacing between images, e. g. ~, \quad, \qquad, \hfill etc.
        \begin{subfigure}[b]{0.30\textwidth}
               \includegraphics[width=\textwidth]{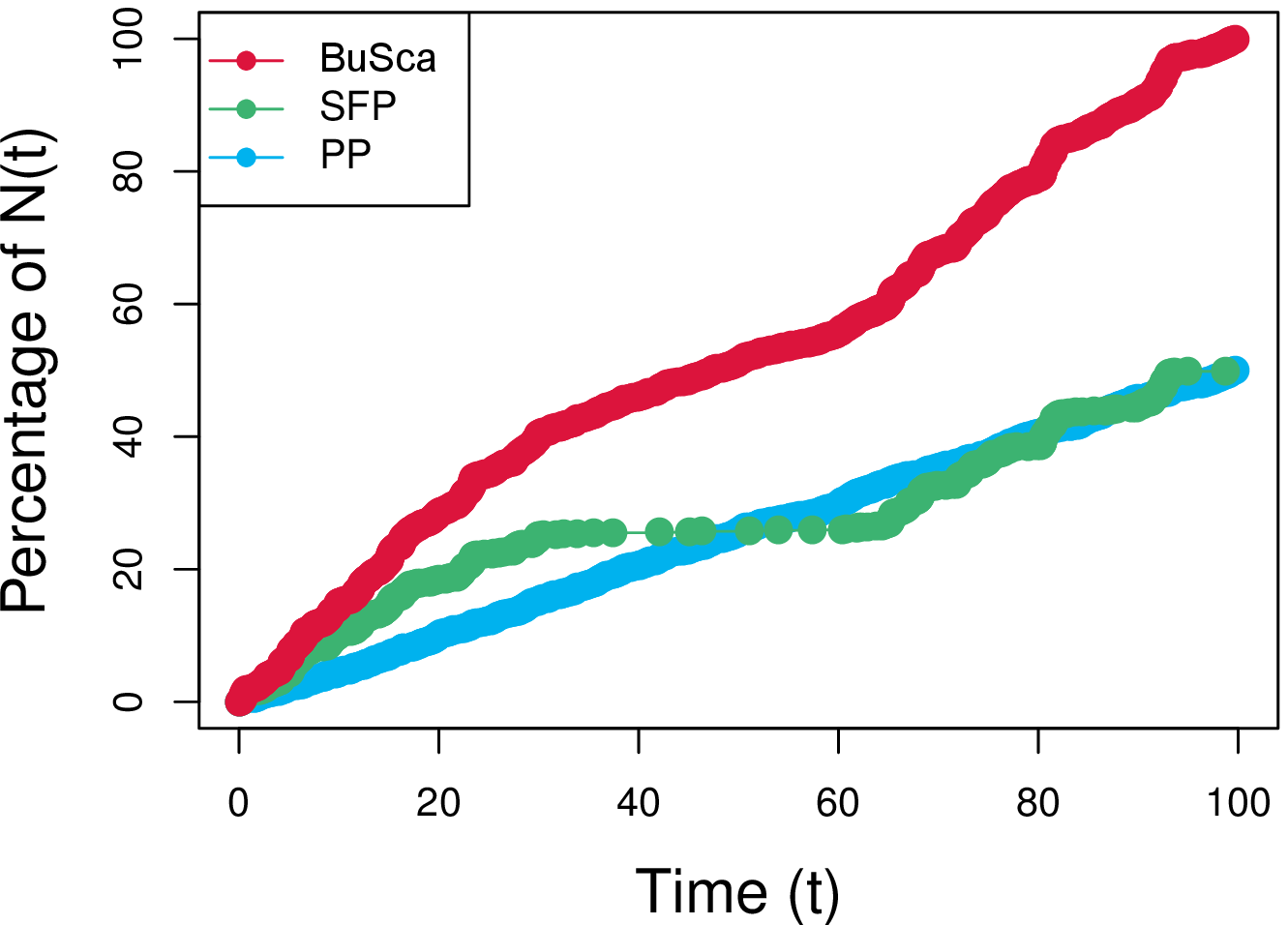}
                \caption{$\paramScale = 50$}
                \label{fig:mix_mu_100_lamb_045_1}
        \end{subfigure}
        ~ %add desired spacing between images, e. g. ~, \quad, \qquad, \hfill etc.
        \begin{subfigure}[b]{0.30\textwidth}
                 \includegraphics[width=\textwidth]{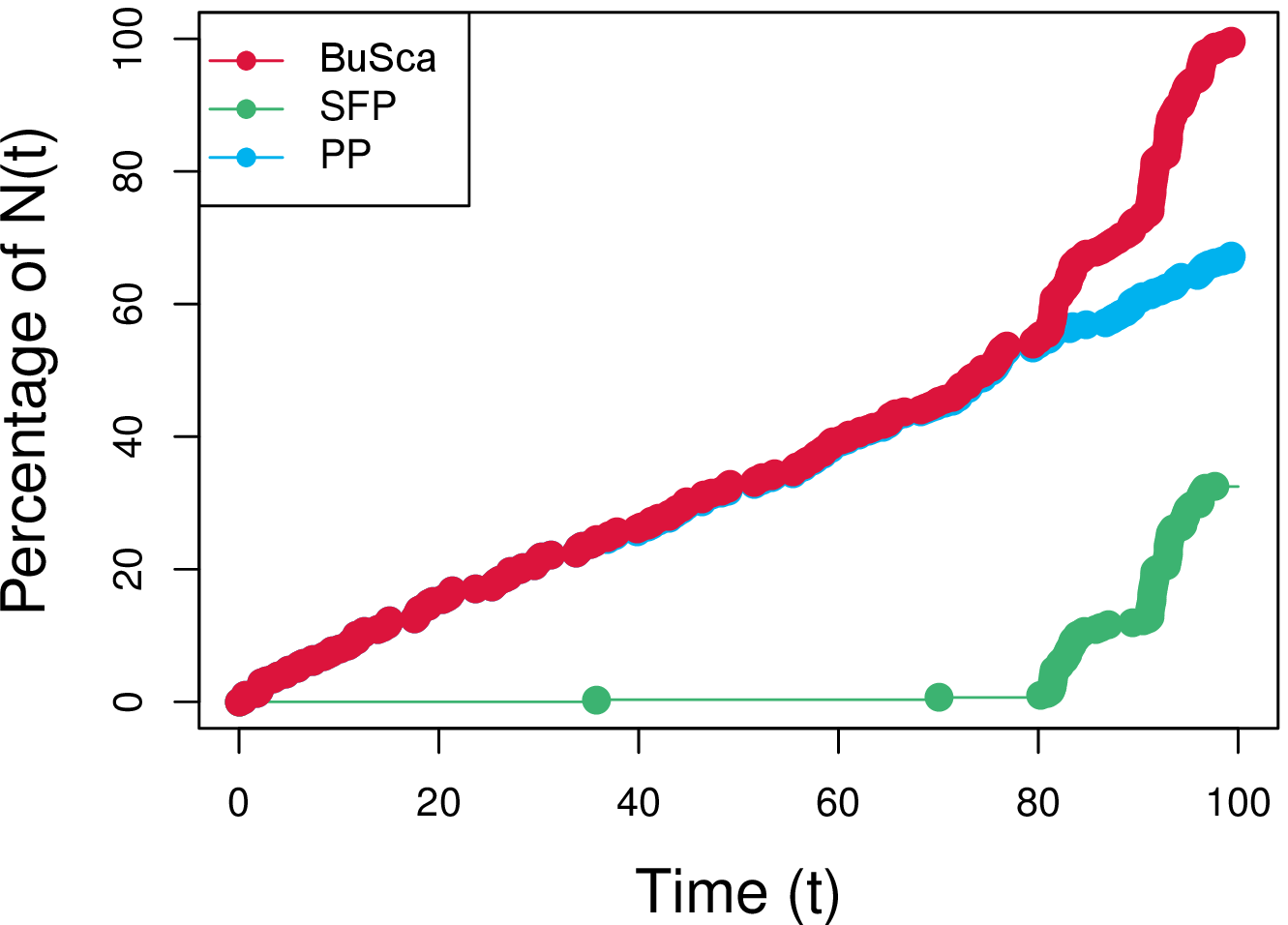}
                \caption{$\paramScale = 25$}
                \label{fig:mix_mu_800_lamb_050_1}
        \end{subfigure}
    \caption{Realizations of the mixture process with different values for $\paramScale$ in the time interval [0,100)}\label{fig:mix_exemplos}
 
\end{figure*}

A point process is called a simple point process if its realizations contain no coincident points. Since both, the SFP and the Poisson
process, are simple, so it is the mixture point process. Also, this guarantees that each event belongs to one of the two component processes.
%We define the \textit{Burstiness scale} parameter as $\paramScale = (1 - \lambda_p/(\lambda_p + (b-a)/\mu))~ 100\%$, which represents the percentage of SFP events expected in the RSE. 
Figure \ref{fig:mix_exemplos} shows different realizations of the mixture process in the time interval $[0,100]$. In each plot, the curves show
the cumulative number of events up to time $t$. The blue line represents a homogeneous Poisson process realization with parameter $\lambda_p$
while the green curve represents the SFP with parameter $\mu$. The red curve represents the mixture of the two other realizations.  

%%%%%%%%%%%%%%%%%%%%%%%%%%%%%%%%%%%%%%%%%%%%%%
\subsection{The likelihood function}
\label{sec_lkd_mix}
%%%%%%%%%%%%%%%%%%%%%%%%%%%%%%%%%%%%%%%%%%%%%%
The log-likelihood function $\ell(\theta)$ for any point process is a function of the conditional intensity $\lambda(t|\mathcal{H}_t)$ 
and of the events $t_1 < t_2 < \ldots$:
\begin{equation}
\begin{array}{rcl}
\ell(\theta) = (\sum_{i=1}^{n}\log{\lambda(t_i|\mathcal{H}_{t_i}))} - \int_a^b  \lambda(t|\mathcal{H}_{t}) dt
\end{array}
\label{eq:like_std_log}
\end{equation}

The conditional intensity function $\lambda_{m}(t|\mathcal{H}_{t})$ of \mixed is the sum of the conditional
intensities of 
the component processes, the Poisson intensity $\lambda_{p}(t|\mathcal{H}_{t}) = \lambda_p$, and the SFP 
intensity $\lambda_s(t|\mathcal{H}_t)$: 
\begin{equation}
\lambda_{m}(t|\mathcal{H}_{t}) = \lambda_p + \lambda_{s}(t|\mathcal{H}_{t}) 
\label{eq:intensidade_mistura}
\end{equation}
The stochastic history $\mathcal{H}_t$ of the mixed process contains only the events' times $t_1, t_2, \ldots$
but not their identifying component processes labels, either $s$ (from SFP) or $p$ (from PP), for each event. 
The log-likelihood function for the mixture process observed in the time interval $[a, b)$ is given by 
\begin{equation}
\ell(\theta) = \sum_{i=1}^{n} \log{\left [ \lambda_{p}+ \lambda_{s}(t_i|\mathcal{H}_{t_i})  \right ]} - \int_a^b   \lambda_{s}(t|\mathcal{H}_{t})  dt - (b-a) \lambda_{p}
\label{eq:like_mix_log}
\end{equation}

The log-likelihood (\ref{eq:like_mix_log}) is not computable because $\lambda_{s}(t|\mathcal{H}_{t})$ requires the knowledge of the last SFP inter-event time 
$\Delta t_i$ for each $t \in (a,b]$. This would be known only if the  source-process label for each event in the observed mixture is also known. 
Since these labels are hidden, we adopt the EM algorithm to obtain the maximum likelihood estimates
$\hat{\lambda}_p$ and $\hat{\mu}$.  
We define the \textit{burstiness scale} $\paramScale = (1 - \lambda_p/(\lambda_p + (b-a)/\mu))~ 100\%$ as the percentage of bursty events in a given RSE. It can be estimated by 
$\hat{\paramScale} = (1 - \hat{\lambda}_p (b-a)/n) ~ 100\%$. This gives the estimated proportion of 
events that comes from the pure SFP process. 
The latent labels are also inferred as part of the inferential procedure. 

The use of the EM algorithm in the case of point process mixtures is new and presents several special challenges with respect to the usual 
EM method. The reason for the difficulty is the lack of independent pieces in the likelihood. 
The correlated sequential data in the likelihood 
brings several complications dealt with in the next two sections.

\subsection{The E step}

The EM algorithm requires the calculation of $\mathbb{E}[\ell(\theta)]$, 
the expected value of the log-likelihood (\ref{eq:like_mix_log}) 
with respect to the hidden labels. 
Since 
\begin{equation*}
\mathbb{E}[\log(X)] \approx \log{ \mathbb{E}[X]} - \frac{\mathbb{V}[X]}{2 \mathbb{E}[X]^2} = 
\log{\mathbb{E}[X]} - \frac{\mathbb{E}[X^2] - \mathbb{E}[X]^2}{2\mathbb{E}[X]^2}
% \label{eq:like_exp_log}
\end{equation*}
we have 
\begin{equation}
\begin{multlined}
\mathbb{E}[\ell(\theta)]\approx 
\sum_{i=1}^{n}  \bigg[ \log{(\lambda_{p} + \mathbb{E}[\lambda_{s}(t_i|\mathcal{H}_{t_i})])}  \\
- \frac{\mathbb{E}[\lambda_{s}(t_i|\mathcal{H}_{t_i})^2] 
- \mathbb{E}[\lambda_{s}(t_i|\mathcal{H}_{t_i})]^2}{2(\lambda_{p} + 
\mathbb{E}[\lambda_{s}(t_i|\mathcal{H}_{t_i})])^2} \bigg]
- \int_a^b  \mathbb{E}[\lambda_{s}(t |\mathcal{H}_{t})]  dt - (b-a) \lambda_{p}
\end{multlined}
\label{eq:like_mix_log_ext}
\end{equation}

A naive way to obtain the required 
\begin{equation}
 \mathbb{E}[ \lambda_{s}(t|\mathcal{H}_{t}) ] = \mathbb{E} \left[ \frac{1}{\mu/e + \Delta t_i}  \right] 
 \label{eq:exp_lik_sfp}
 \end{equation}
 is to consider all possible label assignments to the events and its associated probabilities.
Knowing which events belong to the SFP component, we also know
the value of $\Delta t_i$. Hence, it is trivial to evaluate $1/(\mu/e + \Delta t_i) $ in each one these assignments for any $t$, 
and finally to obtain (\ref{eq:exp_lik_sfp}) by summing up all these values multiplied be the corresponding label assignment probabilities.
This is unfeasible because the number of label assignments is too large, unless the number of events is unrealistically small. 

To overcome this difficulty, we developed a dynamic programming algorithm. 
Figure (\ref{fig:mix_graf_t_i-1_null}) shows the conditional intensity $\lambda_s(t | \mathcal{H}_t)$ up to,
and not including, $t_{i-1}$ as green line segments and the constant Poisson intensity $\lambda_p$ as a blue line.  
Our algorithm is based on a fundamental observation: if $t_{i-1}$ comes from 
the Poisson process, it does not change the current SFP conditional intensity until the next event $t_{i}$ comes in. 

We start by calculating $a_i \equiv \mathbb{E}[\lambda_{s}(t_i|\mathcal{H}_{t_i})] $ conditioning on the 
$t_{i-1}$ event label: 
\begin{equation}
\begin{multlined}
a_i = \mathbb{E} [ \lambda_{s}(t_i|\mathcal{H}_{t_i})] =\\
~~~ \mathbb{P} \left( t_{i-1} \in 
\text{Poisson} | \mathcal{H}_{t_i} \right) 
\mathbb{E} [ \lambda_{s}(t_i|\mathcal{H}_{t_i}, t_{i-1} \in \text{Poisson})]      \\
+ \mathbb{P} \left( t_{i-1} \in \text{SFP} | \mathcal{H}_{t_i} \right) \mathbb{E} [ \lambda_{s}(t_i|\mathcal{H}_{t_i}, t_{i-1} \in \text{SFP})]
\end{multlined}
\label{eq:ai}
\end{equation}

The evaluation of $\lambda_{s}(t_{i} |\mathcal{H}_{t_{i}})$  depends on the label assigned to $t_{i-1}$. If $t_{i-1} \in \text{Poisson}$, as in Figure~\ref{fig:mix_graf_t_i-1_poisson}, the 
last SFP inter-event time interval is the same as that for $t_{i-2} \leq t < t_{i-1}$, since a Poisson event does not change the SFP 
conditional intensity. Therefore, in this case,
\begin{equation}
 \mathbb{E} \left[  \lambda_{s}(t_i|\mathcal{H}_{t_i}, t_{i-1} \in \text{Poisson}) 
 \right] = \mathbb{E} \left[ \lambda_{s}(t_{i-1}|\mathcal{H}_{t_{i-1}} ) \right] = a_{i-1} \: .
 \label{eq:int_lastpoisson}
\end{equation}
For the integral component in (\ref{eq:like_mix_log_ext}), we need the conditional intensity for $t$ in the 
continuous interval and not only at the observed $t_i$ values. However, by the same argument used for $t_i$, we have 
\[ \lambda_{s}(t|\mathcal{H}_{t}, t_{i-1} \in \text{Poisson}) = \lambda_{s}(t_{i-1}|\mathcal{H}_{t_{i-1}} ) \]
for $t \in [t_{i-1}, t_{i})$. 

The probability that the $(i-1)$-th mixture event comes from the SFP component is proportional to its conditional intensity at the event time $t_i$: 
\begin{equation}
 \mathbb{P}\left( t_{i-1} \in \text{SFP} | \mathcal{H}_{t_i} \right) = 
   \frac{\lambda_{s}(t_{i-1}|\mathcal{H}_{t_{i}}) }{\lambda_{s}(t_{i-1}|\mathcal{H}_{t_i}) + \lambda_{p} } \approx 
 \frac{a_{i-1}}{a_{i-1} + \lambda_p} \: . 
\label{eq:pr_sfp}
\end{equation}
Therefore, using (\ref{eq:int_lastpoisson}) and (\ref{eq:pr_sfp}), 
we can rewrite (\ref{eq:ai}) approximately as a recurrence relationship:  
\begin{equation}
a_i = \frac{\lambda_p}{a_{i-1} + \lambda_p} a_{i-1} + \frac{a_{i-1}}{a_{i-1} + \lambda_p} 
\mathbb{E} [ \lambda_{s}(t_i|\mathcal{H}_{t_i}, t_{i-1} \in \text{SFP})] \: . 
\label{eq:esp_lambda02}
\end{equation}

We turn now to explain how to obtain the last term in (\ref{eq:esp_lambda02}). 
If $t_{i-1} \in \text{SFP}$, as exemplified in Figure~\ref{fig:mix_graf_t_i-1_sfp}, 
the last SFP inter-event time must be updated and it will depend on the most recent SFP event previous to $t_{i-1}$.
There are only $i-2$ possibilities for this last previous SFP event 
and this fact is explored in our dynamic programming algorithm. Recursively, we 
condition on these possible $i-2$ possibilities to evaluate the last term in (\ref{eq:esp_lambda02}). 
More specifically, the value of $\mathbb{E} [ \lambda_{s}(t_i|\mathcal{H}_{t_i}, t_{i-1} \in \text{SFP})] $ 
is given by 
\begin{equation}
\begin{multlined}
\mathbb{E} [ \lambda_{s}(t_i|\mathcal{H}_{t_i},(t_{i-1},t_{i-2}) \in \text{SFP}]
   \mathbb{P} (t_{i-2} \in \text{SFP} | \mathcal{H}_{t_i}, t_{i-1} \in \text{SFP})  \\ 
+ \mathbb{E} [ \lambda_{s}(t_i|\mathcal{H}_{t_i}, t_{i-1} \in \text{SFP}, t_{i-2} \in \text{Poisson})] \\
  \mathbb{P} (t_{i-2} \in \text{Poisson} | \mathcal{H}_{t_i}, t_{i-1} \in \text{SFP}) \\
\approx  \frac{1}{\mu/e + (t_{i-1}-t_{i-2})} \frac{a_{i-2}}{a_{i-2}+\lambda_p} ~~~~~~~~~~ \\
+    \mathbb{E} [ \lambda_{s}(t_i|\mathcal{H}_{t_i}, t_{i-1} \in \text{SFP}, t_{i-2} \in \text{Poisson})] 
     \frac{\lambda_p}{a_{i-2}+\lambda_p}
\end{multlined}
\label{eq:recursive01}
\end{equation}
When the last two events $t_{i-1}$ and $t_{i-2}$ come from the SFP process, we know that 
the conditional intensity of the SFP process is given by the first term in (\ref{eq:recursive01}).
The unknown expectation in (\ref{eq:recursive01}) is obtained by conditioning in the $t_{i-3}$ label.
In this way, we recursively walk backwards, always depending on one single unknown of the form 
\[ \mathbb{E} \left[ \lambda_{s}\left( t_i|\mathcal{H}_{t_i}, t_{i-1} \in \text{SFP}, \{ t_{i-2}, t_{i-3}, \ldots, t_{k} \}
 \in \text{Poisson} \right) \right] 
 \]
 where $k < i-2$. At last, we can calculate $a_i$ in (\ref{eq:esp_lambda02}) by the iterative expression 
\begin{equation}
\begin{multlined}
 \mathbb{E} [ \lambda_{s}(t_i|\mathcal{H}_{t_i}, t_{i-1} \in \text{SFP})] =  \\ 
~~~~~~\frac{a_{i-1}}{a_{i-1} + \lambda_p}  \sum_{k=1}^{i-2} \left \{  \frac{a_{k}}{a_{k} + \lambda_p}  
\frac{1}{\mu/e + (t_{i-1} - t_k)}  \prod_{j=k+1}^{i-2}  \frac{\lambda_p}{a_j + \lambda_p} 
\right\}
\label{eq:lamba_s_final_step_e}
\end{multlined}
\end{equation}

We have more than one option as initial conditions for this iterative computation. 
One is to assume that the first two events belong to the SFP. Another one is to use 
\[ \lambda_s \left( t_i| \mathcal{H}_{t_i}, t_{i-1} \in \text{SFP}, 
   \{ t_{i-2}, t_{i-3}, \ldots, t_{2} \}
 \in \text{Poisson} \right) = \frac{1}{\mu/e + \mu} \]
and the first event comes from the SFP. Even with a moderate number of events, this 
initial condition choices affect very little the final results and either of them can be 
selected in any case.

To end the E-step, the log-likelihood in  (\ref{eq:like_mix_log_ext}) requires also 
$\mathbb{E}[\lambda_{s}(t_i|\mathcal{H}_{t_i})^2]$. This is calculated in an entirely analogous way as we did 
above. 

\subsection{The M step}

Different from the E step, the M step did not require special development from us. Having obtained the log-likelihood 
(\ref{eq:like_mix_log_ext}) we simply maximize the likelihood and update the estimated
 parameter values of $\hat{\mu}$ and $\hat{\lambda}_p$. In this maximization procedure, we constrain the 
search within two intervals. For an observed point pattern with $n$ events,  we use $[0, n/{t_n} ]$ for $\lambda_p$. 
The intensity must be positive and hence, the zero lower bound represents a pure SFP process
while the the upper bound represents the maximum likelihood estimate of $\lambda_p$ in the other extreme 
case of a pure homogeneous Poisson process. For the $\mu$ parameter, we adopt the search interval 
$[0, t_n]$. Since $\mu$ is the median inter-event time in the SFP component, a value $\mu \approx 0$ 
induces a pattern with a very large of events while $\mu = t_n$ represents, in practice, a pattern with no 
SFP events. 

\newcommand{\grafinstensidade}{% horizontal axis
\draw(0.65,0) -- (10,0) node[anchor=north] {};
\draw[->] (10.15,0) -- (12,0) node[anchor=west] {$t$};
\draw (0,0) -- (0.5,0) node[anchor=north] {};
% labels
\draw	(0,0) node[anchor=north] {0}

		(1.96,0) node[anchor=north] {$t_{i-7}$}
		(2.54,0) node[anchor=north] {$t_{i-6}$}
		(3.30,0) node[anchor=north] {$t_{i-5}$}
		(4.1,0) node[anchor=north] {$t_{i-4}$}
		(5.62,0) node[anchor=north] {$t_{i-3}$}

		(6.36,0) node[anchor=north] {$t_{i-2}$}
		(7.69,0) node[anchor=north] {$t_{i-1}$}
		( 8.5,0) node[anchor=north] {$t_{i}$}
		(9.11,0) node[anchor=north] {$t_{i+1}$}
		
		(11,0) node[anchor=north] {$t_{n}$};

    \begin{scope} 
      \foreach \x in {0.5,0.65,10,10.15}
 	\draw (\x,-0.25) -- (\x,0.25);

    \end{scope} 

    \begin{scope} 
      \foreach \x in {1.96,2.54,3.30,4.1,5.62,6.36,7.69,8.5}
 	\draw[dotted] (\x,0) -- (\x,1.5);

\foreach \x in {1.96, 4.1,5.62}
            	\filldraw[fill=blue] (\x, 0) circle (.1);
\foreach \x in {2.54,3.30,6.36}
            	\filldraw[fill=green] (\x, 0) circle (.1);

\foreach \x in {7.69, 9.11,11}
            	\filldraw (\x, 0) circle (.1);

	\filldraw[fill=red] (8.5, 0) circle (.1);

    \end{scope} 

% vertical axis
\draw[->] (0,0) -- (0,1.5) node[anchor=south] {$\lambda_{S}(t|H_t)$};
% nominal speed

% Us
\draw[thick] (1,1) -- (12,1) node[right] {$\lambda_{PP}$};
\draw[thick,color=blue] (0,1) -- (12,1) ;

\draw[thick,color=green] (0,1.05) -- (0.5,1.05) ;
\draw[thick,color=green] (0.65,0.8) -- (2.54,0.8) ;
\draw[thick,color=green] (2.54,0.58) -- (3.30,0.58) ;
\draw[thick,color=green] (3.30,1.2) -- (6.36,1.2) ;
\draw[thick,color=green] (6.36,0.3) -- (7.69,0.3) ;
}

\begin{figure}
\centering
\begin{tikzpicture}[scale=0.65, every node/.style={scale=0.65}]
		\grafinstensidade
		\end{tikzpicture}
	\caption{Start}
      \label{fig:mix_graf_t_i-1_null}
\end{figure}

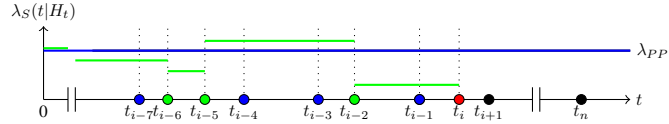
\begin{figure}
\centering
\begin{tikzpicture}[scale=0.65, every node/.style={scale=0.65}]
		\grafinstensidade
		\filldraw[fill=blue] (7.69, 0) circle (.1);
		\draw[thick,color=green] (7.69,0.3) -- (8.5,0.3);
		\end{tikzpicture}
                \caption{$t_{i-1} \in Poisson$}
                \label{fig:mix_graf_t_i-1_poisson}
\end{figure}
\begin{figure}

\centering
  \begin{tikzpicture}[scale=0.65, every node/.style={scale=0.65}]
		\grafinstensidade
		\filldraw[fill=green] (7.69, 0) circle (.1);
		\draw[thick,color=green] (7.69,0.66) -- (8.5,0.66);
		\end{tikzpicture}
                \caption{$t_{i-1} \in SFP$}
                \label{fig:mix_graf_t_i-1_sfp}
\end{figure}
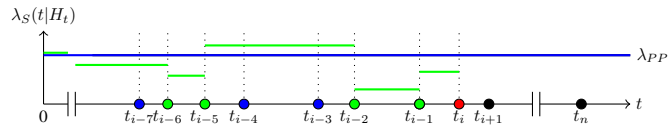

\subsection{Complexity analysis}

The E step calculation is represented by (\ref{eq:like_mix_log_ext}) and it has complexity $O(n^3)$, where $n$ is the number of events. 
The needed $a_i = \mathbb{E} [ \lambda_{s}(t_i|\mathcal{H}_{t_i})]$ 
in (\ref{eq:lamba_s_final_step_e}) requires $O(i^2)$ operations due to the product with $O(i)$ factors which, when summed, will end up with $O(i^2)$. However, we simplify this calculation when we consider only the last 10 iterations of the product operator. It was possible because the sequential multiplication of probabilities reduces significantly the weight of the first events in the calculation of SFP intensity function. The results were very close to 
the non-simplified calculation but with $O(10i)$ operations.
After the $a_i$ are calculated, the integral in (\ref{eq:like_mix_log_ext})  is simply the evaluation of the area under a step function with $n$ steps and therefore needs $O(n)$ calculations. Substituting the terms in (\ref{eq:like_mix_log_ext})  by their complexity order and summing them, we find that the E step requires $O(n^2)$ operations. 

In the M step, the main cost is related to the number of runs that the maximization algorithm requires. We used coordinate ascent for each parameter $\mu$ and $\lambda_p$.
The EM steps are repeated until convergence or a maximum of $m$ steps is reached. Therefore, the final complexity of our algorithm is $O(mn^2)$.
In our case, we used $m=100$ but, on average, the EM algorithm required 7.10 loops considering all real datasets described in Section \ref{sec:dados}.
In 95\% of the cases, it took less than 21 loops. 

%%%%%%%%%%%%%%%%%%%%%%%%%%%%%%%%%%%%%%%%%%%%%%
\section{MLE bias and a remedy}
\label{sec:mlebias}
%%%%%%%%%%%%%%%%%%%%%%%%%%%%%%%%%%%%%%%%%%%%%%

Several simulations were performed to verify that the estimation of the parameters proposed in Section \ref{sec:ourmodel} is suitable. There is no theory about the MLE behavior in the case of point processes data following a complex mixture model as ours. For this, synthetic data were generated by varying the sample size $n$ and the parameters $\lambda_{p}$ and $\mu$ of the mixture. We vary $n$ in $\{100, 200, \ldots, 1000\}$ for each pair $(\lambda_p, \mu)$. 
The parameters $\lambda_p$ and $\mu$ were empirically selected in such a way that the 
expected percentage of points coming from the SFP process (denoted by the burstiness 
scale parameter $\paramScale$) varied in 
$\{10\%, 20\%, \ldots, 90\%\}$. 

For each pair $(n, \paramScale)$, we conducted 100 simulations, totaling 9,000 simulations. 
In each simulation, the estimated parameters $(\hat{\lambda}_p, \hat{\mu})$ were calculated 
by the EM algorithm. Since their range vary along the simulations, we 
considered their relative differences with respect to the true values 
$(\lambda_p, \mu)$. For $\mu$, define 
\begin{equation}
\begin{array}{rcl}

\Delta (\hat{\mu}, \mu) = \left\{
  \begin{array}{l l}
   \hat{\mu}/\mu - 1, & \quad \text{if $\hat{\mu}/\mu \geq 1$ }\\
   1 -  \mu/\hat{\mu},  & \quad \text{otherwise}
  \end{array} \right.
\end{array}
\label{eq:delta_estrela}
\end{equation}
The main objective of this measure is to treat symmetrically the relative differences
between $\hat{\mu}$ and $\mu$.  
Consider a situation where $\Delta (\hat{\mu}, \mu) = 0.5$.This means that 
$\hat{\mu} = 1.5 \mu$. Symmetrically, if $\mu = 1.5 \hat{\mu}$, we have 
$\Delta (\hat{\mu}, \mu) = -0.5$. The value $\Delta (\hat{\mu}, \mu) = 0$ implies
that  $\hat{\mu} =\mu$. We define $\Delta (\hat{\lambda}_p, \lambda_p)$ analogously. 
 
%%%%%%%%%%%%%%%%%%%%%%%%%%%%%%%%%%%%%%%%%%%%%%
\subsection{The estimator $\hat{\lambda}_{p}$}
\label{subsec:lambdaEMSec}
%%%%%%%%%%%%%%%%%%%%%%%%%%%%%%%%%%%%%%%%%%%%%%

\begin{figure}
\centering
\includegraphics[width=0.75\textwidth]{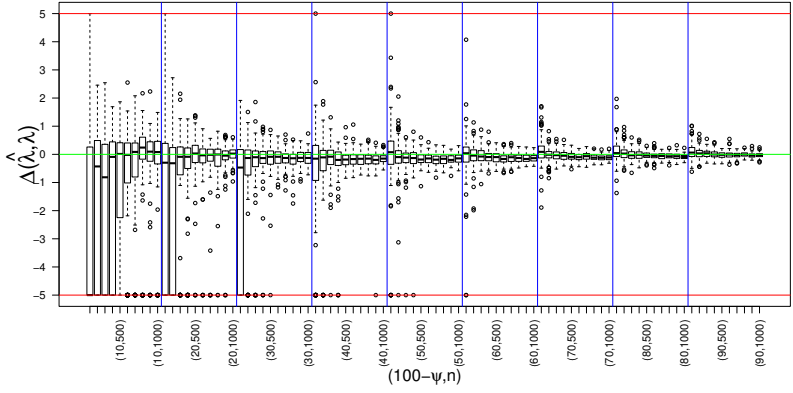}  
         \caption{Boxplots of $\Delta (\hat{\lambda}_p, \lambda_p)$ according to 
         $(100-\paramScale)$ and $n$.}
         \label{fig:teste_lambda_EM}
\end{figure}

The results of $\Delta (\hat{\lambda}_p, \lambda_p)$ are shown in Figure \ref{fig:teste_lambda_EM}. Each boxplot correspond to one of the 90 possible 
combinations of $(\paramScale, n)$. The vertical blue lines separate out the different 
values of $\paramScale$. Hence, the first 10 boxplots are those calculated to 
the combinations $(\paramScale = 90\%, n=100, 200, \ldots, 1000)$.
The next 10 boxplots correspond to the values of $\Delta (\hat{\lambda}_p, \lambda_p)$
for $\paramScale = 80\%$ and $n=100, 200, \ldots, 1000$. The absolute 
values for $\Delta (\hat{\lambda}_p, \lambda_p)$ were censored at $5$ and 
this is represented by the horizontal red lines at heights $-5$ and $+5$.

The estimator $\hat{\lambda}_p$ is well behaved, with small bias and 
variance decreasing with the sample size. The only cases where it has a large 
variance is when the total sample size is very small and, at the same time, the 
percentage of Poisson events is also very small. For example, with $n=200$ and 
$\paramScale = 90\%$, we expect to have only 20 \emph{unidentified} Poisson cases and 
it is not reasonable to expect an accurate estimate in this situation. 

%%%%%%%%%%%%%%%%%%%%%%%%%%%%%%%%%%%%%%%%%%%%%%
\subsection{The estimador $\hat{\mu}$}
\label{subsec:muEMSec}
%%%%%%%%%%%%%%%%%%%%%%%%%%%%%%%%%%%%%%%%%%%%%%

The  results of $\Delta (\hat{\mu}, \mu)$ from the simulations are shown 
in Figure \ref{fig:teste_mu_EM}, in the 
same grouping format as in Figure \ref{fig:teste_lambda_EM}.
It is clear that $\hat{\mu}$ overestimates the true value of $\mu$ in all cases
with the bias increasing with the increase of the Poisson process share.  
In the extreme situation when $\paramScale \leq 20\%$, the large $\hat{\mu}$ leads to 
an erroneous small expected number of SFP events in the observation time interval. 
Indeed, a mixture with Poisson process events only has $\mu = \infty$. 
Additionally to the bias problem, the estimator  $\hat{\mu}$ also has a 
large variance when the SFP process has a small number of events. 

We believe that the poor performance of the EM algorithm estimator $\hat{\mu}$ 
is related to the calculation of the expected value of the likelihood function. This calculation 
was done using approximations to deal with the unknown events' labels,
which directly influences the calculus of the SFP stochastic intensity function. 
This influence has less impact for the $\lambda_p$, since the Poisson process 
intensity is deterministic and fixed during the entire interval.

\begin{figure}
\centering
 \includegraphics[width=0.75\textwidth]{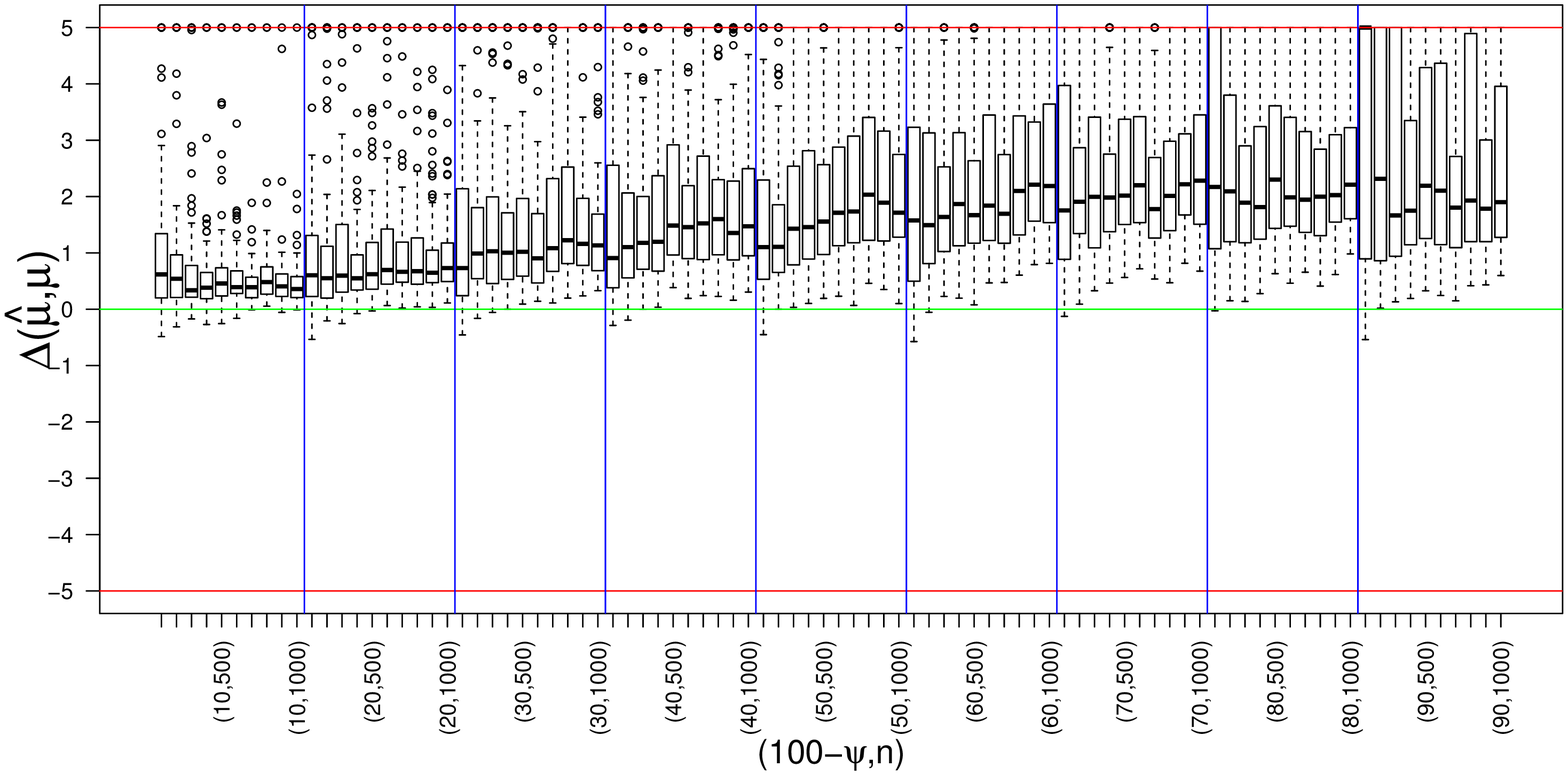}  
          \caption{Boxplots of $\Delta (\hat{\mu}, \mu)$ for the usual MLE $\hat{\mu}$ according to 
         $(100-\psi)$ and $n$.}
         \label{fig:teste_mu_EM}
\centering
\includegraphics[width=0.75\textwidth]{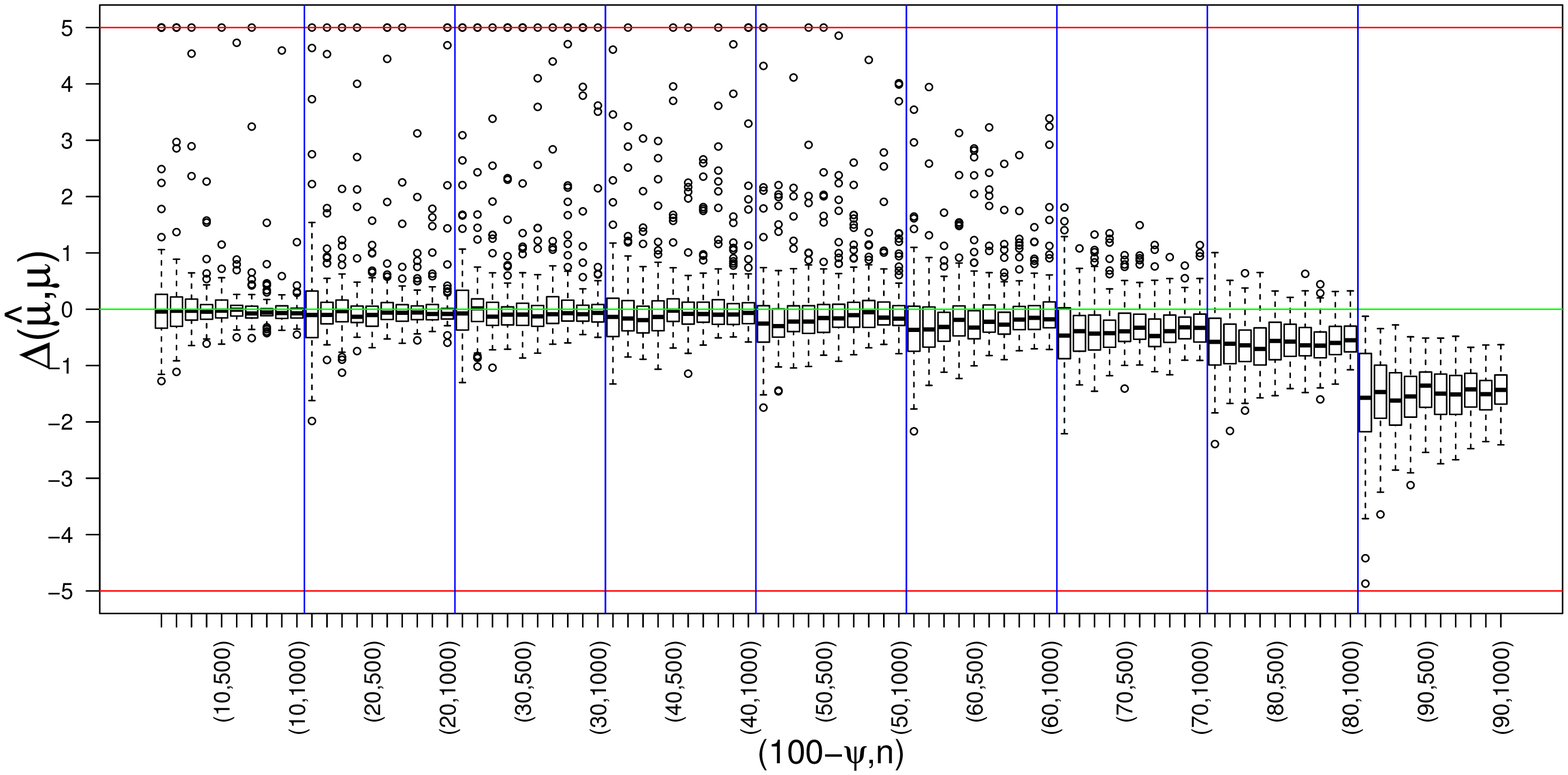}  
     \caption{Boxplots of $\Delta (\hat{\mu}, \mu)$ for the improved estimator $\hat{\mu}$ according to 
         $(100-\psi)$ and $n$.}
         \label{fig:teste_mu_mediana}
\end{figure}

As $\mu$ is the median inter-event time in a pure SFP process, 
a simple and robust estimator in this pure SFP situation 
is the empirical median of the intervals $t_{i+1}-t_{i}$. 
Our alternative estimator for $\mu$ deletes some carefully selected 
events from the mixture, reducing the dataset to a pure SFP process and, then, taking 
the median of the inter-event times of the remaining events. More specifically, 
conditioned on the well behaved $\hat{\lambda}_p$ estimate, we generate pseudo events 
$u_1 < u_2 < \ldots < u_m$ coming from a homogeneous Poisson process and within 
the time interval $(0,t_n)$. Sequentially define 
\[ t^*_k = \arg \min_{t_i \in S_k} |u_k - t_i| \]
with $S_k = T_k \cap R_k$ where 
\[ T_k = \{ t_1, t_2, \ldots, t_n \} - \{ t^*_1, \ldots, t^*_{k-1} \}
\]
and $R_k = \{ t_i : |u_k - t_i| < 2/\lambda_p \}$. 
This last constraint avoids the deletion to be entirely concentrated in bursty regions.
We assume that the left over events in $T_m$ constitute a realization of a 
pure SFP process and we use their median inter-event time as an 
estimator of $\mu$. As this is clearly 
affected by the randomly deleted events $t^*_k$, we repeat this 
procedure many times and average the results to end up with a final estimate, 
which we will denote by  $\hat{\mu}$. 

The results obtained with the new estimator of $\mu$ can be visualized in Figure 
\ref{fig:teste_mu_mediana}. Its estimation error is clearly smaller than that 
obtained directly by the EM algorithm, with an underestimation of $\mu$ only
when the Poisson process component is dominant. This is expected because when the SFP 
component has a small percentage of events, its corresponding 
estimate is highly variable. In this case, there will be a large number of supposedly 
Poisson points deleted, remaining few SFP events to estimate the $\mu$ parameter,
implying a high instability. 

\begin{figure}
\centering
\includegraphics[width=0.75\textwidth]{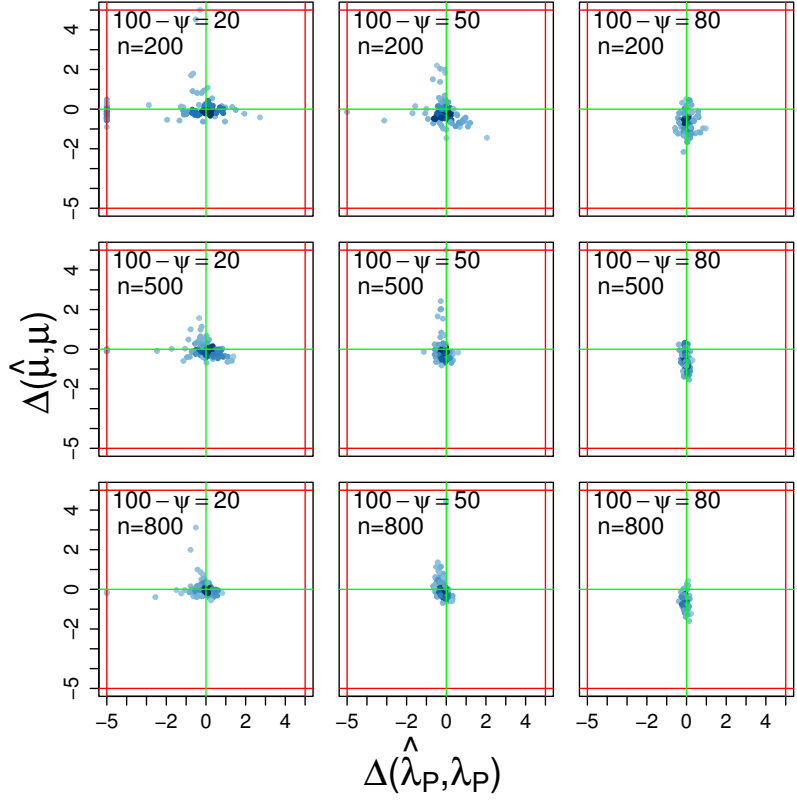}  
         \caption{$ \Delta (\hat{\lambda}_p, \lambda_p)$  versus $\Delta (\hat{\mu}, \mu)$.}
         \label{fig:mu_med_versus_lambda_EM}
\end{figure}

In Figure \ref{fig:mu_med_versus_lambda_EM} we can see the two estimation errors simultaneously. Each dot represents one pair 
$\left( \Delta (\hat{\mu}, \mu), \Delta (\hat{\lambda}_p, \lambda_p) \right)$. They are concentrated around the origin and do not show any 
trend or correlation. This means that the estimation errors are approximately independent of each other and that the estimates are close to their 
real values. 

%%%%%%%%%%%%%%%%%%%%%%%%%%%%%%%%%%%%%%%%%%%%%%
\section{Classification test} 
\label{sec:classtest}
%%%%%%%%%%%%%%%%%%%%%%%%%%%%%%%%%%%%%%%%%%%%%%

When analysing a point process dataset, a preliminary analysis should test if a simpler point process, comprised either by a pure SFP or a 
pure Poisson process, fits the observed data as well as the more complex mixture model. Let $\theta = (\lambda_p, \mu)$ and the 
unconstrained parameter space be $\Theta = [0,\infty]^2$. We used the maximum likelihood ratio test statistic $R$
of $H_1: \theta \in \Theta$ against the null hypothesis $H_0: \theta \in \Theta_0$ where, alternatively, we consider either
$\theta \in \Theta_0 = (0, \infty) \times \{ \infty \}$ or  $\theta \in \Theta_0 = \{ 0 \} \times (0, \infty)$ to represent the 
pure Poisson and the pure SFP processes, respectively. Then 
\[ R = 2 \times \left( \max_{\theta \in \Theta} \ell(\theta) - \max_{\theta \in \Theta_0} \ell(\theta_0) \right) 
\]
where the log-likelihood $\ell(\theta)$ is given in (\ref{eq:like_std_log}). As a guide, we used a threshold $\alpha = 0.05$ to deem the test 
significant. As a practical issue, since taking the median inter-event time $\mu$ of the SFP process equal to $\infty$ is not 
numerically feasible, we set it equal to the length of the observed total time interval.  

As there is one free parameter in each case, one could expect that the usual asymptotic distribution of $2\log(R)$ should follow a chi-square 
distribution with one degree of freedom. However, this classic result requires several strict assumptions about the stochastic nature of the data, 
foremost the independence of the observations, which is not the situation in our model.
Therefore, to check the accuracy of this asymptotic distribution to gauge the test-based decisions, we carried out 2000 additional Monte Carlo simulations,
half of them following a pure Poisson process, the other half following a pure SFP. Adding these pure cases to those of the mixed cases at different 
percentage compositions described previously, we calculated the test p-values $\phi_{p}$ and $\phi_s$ based on the usual chi-square distribution 
with one degree of freedom. Namely, with $\mathbb{F}$ being the cumulative distribution function of the chi-square distribution 
with one degree of freedom, we have 
\begin{equation}
\phi_p = 1 - \mathbb{F}(R) =   1 -  \mathbb{F}\left(2 \max_{\lambda_p, \mu} \ell(\lambda_p, \mu)  - 2 \max_{\lambda_p} \ell(\lambda_p, \infty)  \right)
\label{eq:prpoisson_teste}
\end{equation}
and 
\begin{equation}
\phi_s = 1 - \mathbb{F}(R) =   1 -  \mathbb{F}\left(2 \max_{\lambda_p, \mu} \ell(\lambda_p, \mu)  - 2 \max_{\mu} \ell(0, \mu)  \right) \: . 
\label{eq:prSFP_teste}
\end{equation}

\begin{figure}
\centering
\includegraphics[width=0.75\textwidth]{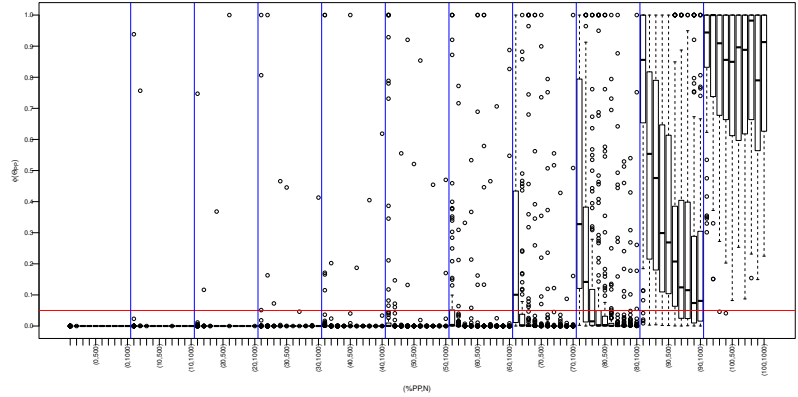}  
         \caption{$H_0: (\lambda_p, \mu) \in \Theta_0 = (0, \infty) \times \{ \infty \}$, the pure Poisson process.}
         \label{fig:teste_hip_poisson}
\end{figure}

\begin{figure}
\centering
\includegraphics[width=0.75\textwidth]{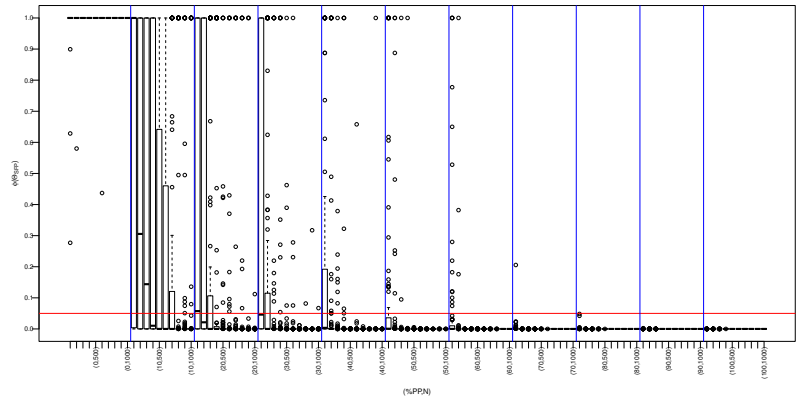}  
         \caption{$H_0: (\lambda_p, \mu) \in \Theta_0 = \{ 0 \} \times (0, \infty)$, the 
         pure SFP process.}
         \label{fig:teste_hip_SFP}
\end{figure}

The p-values $\phi_p$ and $\phi_s$ of all simulated point processes, pure or mixed, 
can be seen in the boxplots of Figures \ref{fig:teste_hip_poisson} and \ref{fig:teste_hip_SFP}, respectively. 
The red horizontal lines represent the 0.05 threshold. Considering initially the 
plots in Figure \ref{fig:teste_hip_poisson}, the first block of 10 boxplots correspond to 
a pure SFP (when $\paramScale = 100$) with different number of events. The p-values are practically 
collapsed to zero and the test will reject the null hypothesis that the process is a pure Poisson process, which is the correct decision. Indeed, this correct decision is taken in virtually all 
cases until $\paramScale \geq 40\%$. The test still correctly rejects the pure Poisson in all cases 
where $\paramScale > 20\%$ except when the number of events is very small. Only when the $\paramScale = 10\%$
or $\paramScale = 0\%$ (and, therefore, it is pure Poisson process) the p-value distribution clearly shifts upward and starts accepting the null hypothesis frequently. This is exactly the expected and desired behavior for our test statistic. Figure \ref{fig:teste_hip_SFP} presents the behavior of 
$\phi_s$ and its behavior is identical to that of $\phi_p$. 

\begin{figure}
\centering
	      \includegraphics[width=0.50\textwidth]{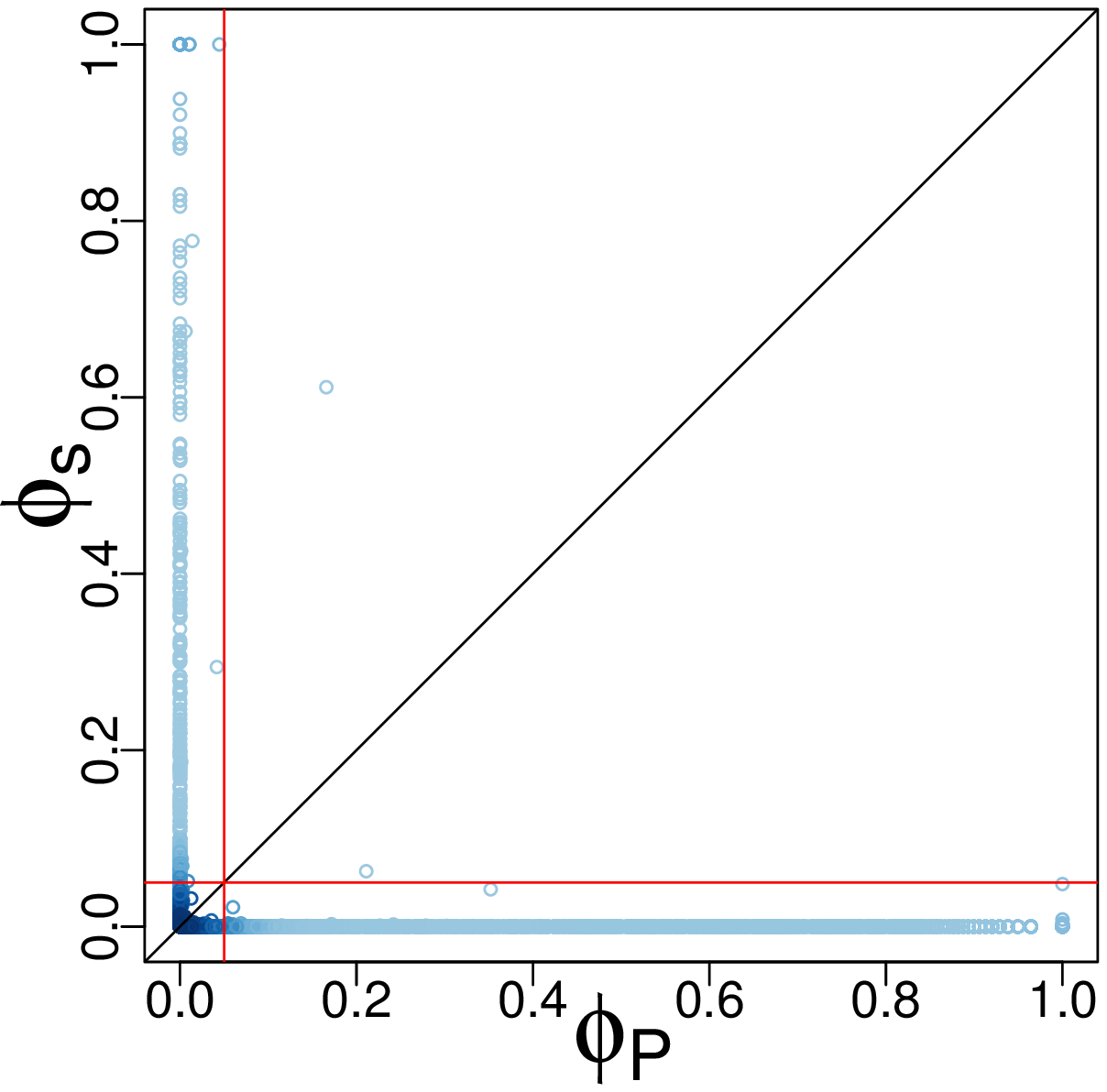}  
        \caption{$\phi_p$ versus $\phi_s$}
        \label{fig:ppoissonXpsfp}
\end{figure}

Figure \ref{fig:ppoissonXpsfp} shows the joint behavior of $(\phi_p, \phi_s)$ The red vertical and horizontal lines represent the $0.05$ level threshold. Clearly, the two tests practically never 
accept both null hypothesis, the pure Poisson and pure SFP processes. Either one or other 
pure process is accepted or else both pure processes are rejected, indicating a mixed process. 

\section{Parsimonious Characterization}
\label{sec:dados}
%%%%%%%%%%%%%%%%%%%%%%%%%%%%%%%%%%%%%%%%%%%%%%

We used eight datasets split into three groups. The first one contains the comments on topics of 
several web services: the discussion forums \textit{AskMe}, \textit{MetaFilter}, and 
\textit{MetaTalk} and the collaborative recommendation systems \textit{Digg}  and \textit{Reddit}. 
The second group contains user communication events: e-mail exchange (\textit{Enron})  and
hashtag-based chat (\textit{Twitter}). The fourth group is composed by user reviews and 
recommendations of restaurants in a collaborative platform (\textit{Yelp}). In total, we analysed $18,685,678$ events.

The \textit{AskMe}, \textit{MetaFilter}\footnote{\url{http://stuff.metafilter.com/infodump/}- Accessed in September, 2013} and \textit{MetaTalk} datasets were made available 
by the \textit{Metafilter Infodump Project2}. The \textit{Digg}\footnote{\url{http://www.infochimps.com/datasets/diggcom-data-set} - Accessed in September, 2013} dataset was temporarily available 
in the web and was downloaded by the authors. The \textit{Enron}\footnote{\url{https://www.cs.cmu.edu/~./enron/} - Accessed in September, 2013} data were obtained 
through the \textit{CALO Project (A Cognitive Assistant that Learns and Organizes)} of 
Carnegie Mellon University. The \textit{Yelp}\footnote{\url{http://www.yelp.com/dataset_challenge} - Accessed in August, 2014} data were available during the 
\textit{Yelp Dataset Challenge}. 
The \textit{Reddit} and \textit{Twitter} datasets were collected using their respective APIs.
All datasets have time scale where the unit is the second except for Yelp, which has the time scale 
measured in days, a more natural scale for this kind of evaluation review service.  

For all databases, each RSE is a sequence of events timestamps and the event varies according to the dataset.
For the \textit{Enron} dataset, the RSE is associated with individual users and 
the events are the incoming and outgoing e-mail timestamps. For the \textit{Twitter} dataset, each RSE  
is associated with a hashtag and the events are the tweet timestamps mentioning that hashtag.
For the \textit{Yelp} dataset, each RSE is associated with a venue and the events are the 
reviews timestamps. For all other datasets, the RSE is a discussion topic and the events are composed by 
comments timestamps. As verified by \cite{Wang2012}, the rate at which comments arrive has a drastic decay
after the topic leaves the forum main page. The average percentage of comments made before this inflection 
point varies from 85\% to 95\% and these represents the bulk of the topic life. As a safe cutoff point, we 
considered the 75\% of the initial flow of comments in each forum topic. 

Table \ref{tab:datasets} shows the number of RSEs in each database. 
It also shows the average number of events by dataset, as well as the minimum and maximum 
number of events. We applied our classification test from Section \ref{sec:classtest}
and the table shows the percentage categorized as pure Poisson process,
pure SFP, or mixed process. For all datasets, the p-values $\phi_p$ and $\phi_s$ have a 
behavior similar to that shown in Figure \ref{fig:ppoissonXpsfp}, leading us to 
believe in the efficacy of our classification test to separate out the models 
in real databases in addition to their excellent performance in the synthetic databases.
% ?? a result that can be verified in the extended version of this paper in the Arxiv ??

\begin{figure}
\centering
\includegraphics[width=0.75\textwidth]{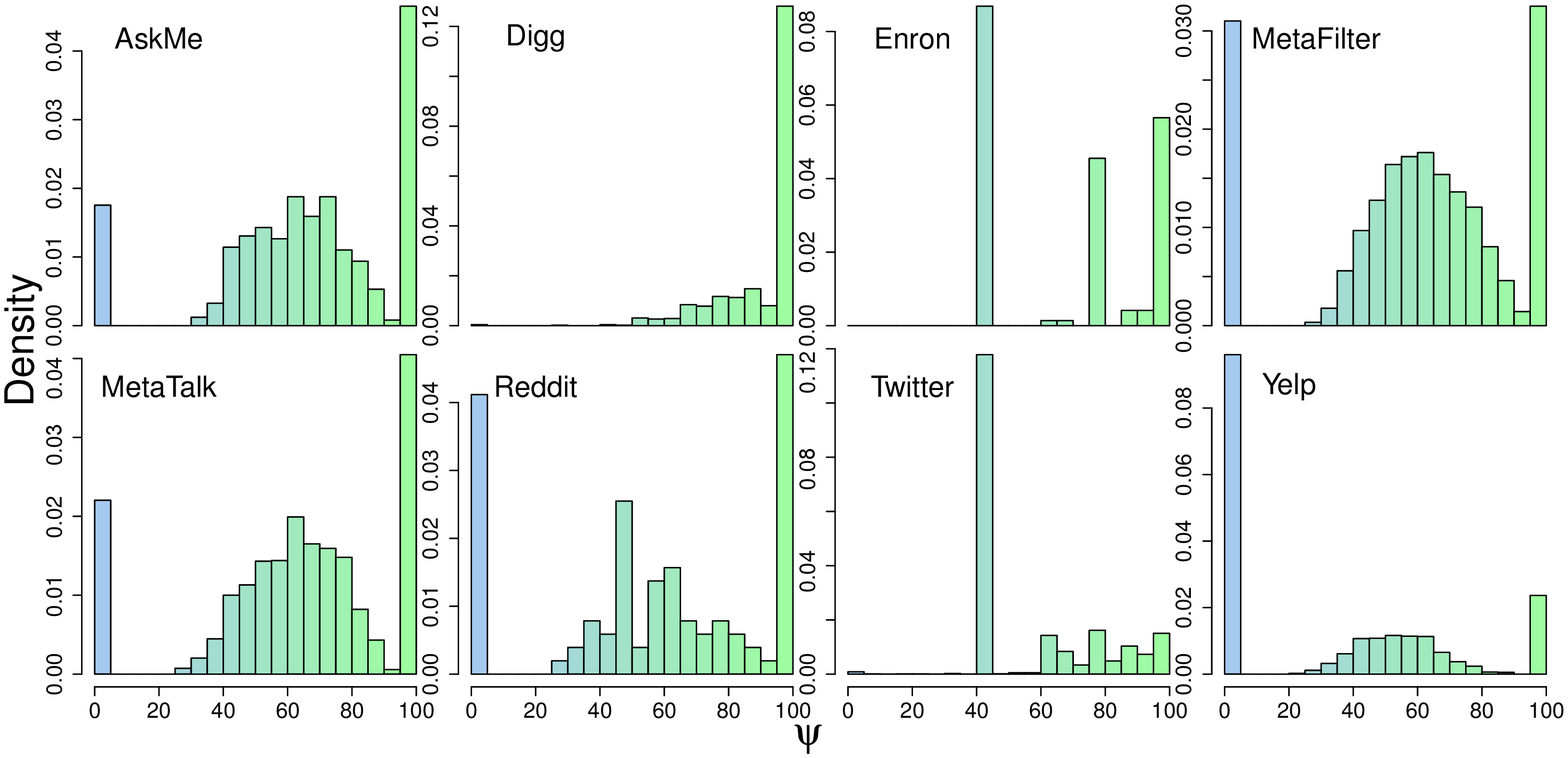}  
         \caption{$\paramScale$ in each dataset.}
         \label{fig:PropPoisson}
\end{figure}

A more visual and complete way to look at the \textit{burstiness scale}  $\paramScale$ is in the histograms of Figure \ref{fig:PropPoisson}. 
The horizontal axis shows the expected percentage of the Poisson process component in the RSE
given by $\hat{\lambda}_p/n$. 
The two extreme bars at the horizontal axis, at $\paramScale = 100$ and $\paramScale = 0$, have areas equal 
to the percentage of series classified as pure SFP and as pure Poisson, respectively. 
The middle bars represent the RSE classified as mixed point processes.
AskMe, MetaFilter, MetaTalk, Reddit, Yelp have the composition where the three models, 
the two pure and the mixed one, appear with substantial amount. The Poisson process 
share of the mixed processes distributed over a large range, from close to zero to large 
percentages, reflecting the wide variety of series behavior. 

\begin{table*}
\centering
\caption{Description of the databases: 
number of series of events; minimum, average, and maximum number of events; classification test results; 
gaussian fit parameters}
\label{tab:datasets}
\resizebox{\textwidth}{!}{%
\begin{tabular}{|c|c|c|c|c|c|c|c|c|c|c|}
\hline
\multirow{2}{*}{\textbf{Base}} & \textbf{\# of} & \multicolumn{3}{c|}{\textbf{\# of events per series}} & \multicolumn{3}{c|}{\textbf{Hypothesis Test}} & \multicolumn{3}{c|}{\textbf{Bivariate Gaussian}} \\ 
\cline{3-11}        &   \textbf{series}  & \textbf{Min}         & \textbf{Avg}        & \textbf{Max}        & \textbf{Mix}         & \textbf{PP}        & \textbf{SFP}         & \textbf{$\overline{\log\lambda_P} ({\sigma^2}_{\log\lambda_P})$}        & \textbf{$\overline{\log\mu} ({\sigma^2}_{\log\mu})$}         & \textbf{$\rho_{(\log\lambda_P,\log\mu)}$}         \\ \hline
AskMe&490&74&99.30&699&333 (67.96\%)&43 (8.78\%)&114 (23.26\%)& -6.91  ( 0.98 )  & 4.77  ( 0.39 )  & -0.40 \\ \hline
Digg&974&39&90.41&296&353 (36.24\%)&2 (0.21\%)&619 (63.55\%)& -8.08  ( 0.83 )  & 4.4  ( 0.38 )  & -0.11 \\ \hline
Enron&145&55&1,541.35&14258&106 (73.1\%)&0 (0\%)&39 (26.9\%)& -11.56  ( 0.62 )  & 8.18  ( 0.57 )  & -0.28 \\ \hline
MetaFilter&8243&72&131.10&4148&5625 (68.24\%)&1279 (15.52\%)&1339 (16.24\%)& -6.76  ( 0.94 )  & 4.78  ( 0.37 )  & -0.39 \\ \hline
MetaTalk&2460&73&151.92&2714&1691 (68.74\%)&271 (11.02\%)&498 (20.24\%)& -7.31  ( 1.08 )  & 5.23  ( 0.55 )  & -0.57 \\ \hline
Reddit&102&37&535.43&4706&58 (56.86\%)&21 (20.59\%)&23 (22.55\%)& -6.12  ( 1.1 )  & 3.26  ( 3.48 )  & -0.85 \\ \hline
Twitter&17088&50&969.68&8564&15913 (93.12\%)&72 (0.42\%)&1103 (6.46\%)& -10.01  ( 0.31 )  & 6.82  ( 5.26 )  & -0.66 \\ \hline
Yelp&1929&50&127.84&1646&774 (40.12\%)&927 (48.06\%)&228 (11.82\%)& -3.79  ( 0.38 )  & 2.28  ( 0.38 )  & -0.22 \\ \hline
\end{tabular}}
\end{table*}

 %%%%%%%%%%%%%%%%%%%%%%%%%%%%%%%%%%%%%%%%%%%%%%
%\subsection{Characterizing parsimoniously the databases}
%\label{comportamentoconjunto}
%%%%%%%%%%%%%%%%%%%%%%%%%%%%%%%%%%%%%%%%%%%%%%

Figure \ref{fig:teste_mu_EM} shows the estimated pairs $(\log\hat{\lambda}_{p}, \log \hat{\mu})$ of each events
stream classified as a mixed process. The logarithmic scale provides the correct scale to 
fit the asymptotic bivariate Gaussian distribution of the maximum likelihood estimator. 
Each point represents a RSE and they are colored according to the database name. Except by the 
Twitter dataset, all others have their estimator distribution approximately fitted by a bivariate 
Gaussian distribution with marginal mean, variance and correlation given in Table \ref{tab:datasets}.

The correlation is negative in all databases, meaning that a large value of the Poisson process parameter 
(that is, a large $\hat{\lambda}_{p}$) tends to be followed by small values of the SFP component 
(that is, a small $\hat{\mu}$, implying a short median inter-event time between the SFP events). 
Not only each database has a negative correlation between the mixture parameters, they also 
occupies a distinct region along a NorthWest-SouthEast gradient. Starting from the upper left corner,
we have the Enron email cloud, exhibiting a low average $\hat{\lambda}_{p}$ and a jointly high 
$\hat{\mu}$. Descending the gradient, we find the less compactly shaped Twitter point could. 
In the $(-8, -6) \times (4, 6)$ region we find the foruns (AskMe, MetaFilter, MetaTalk). Slightly 
shifted to the left and further below (within the $(-10, -7) \times (3, 5)$ region), we find the two
collaborative recommendation systems (Reddit and Digg). Finally, in the lower right corner, we 
have the Yelp random series estimates. 

In this way, our model has been able to spread out the 
different databases in the space composed by the two component processes parameters. 
Different communication services lives in a distinctive location in this mathematical
geography. 

\begin{figure}
\centering
 \includegraphics[width=0.75\textwidth]{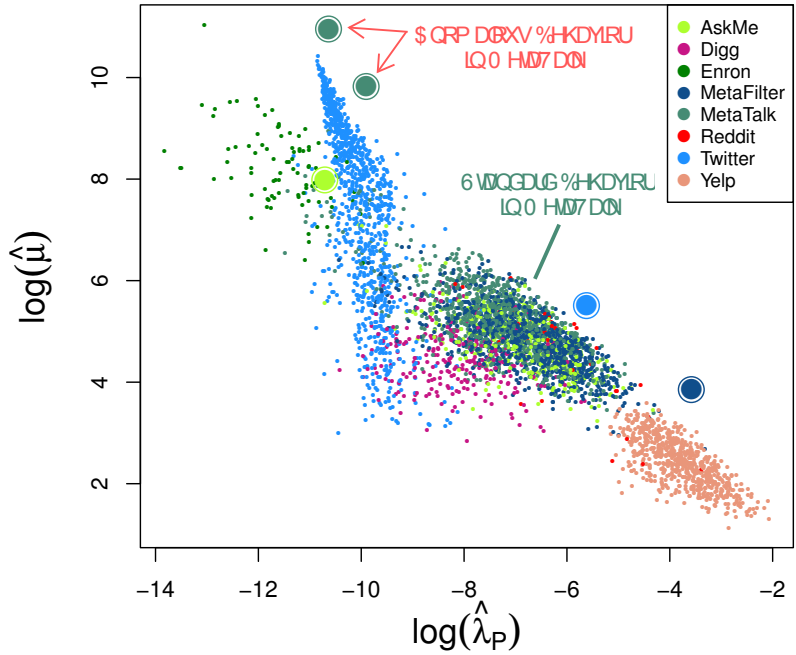}  
          \caption{Estimates $(\hat{\lambda}_{p}, \hat{\mu})$ for all events streams from the eight databases
          (logarithmic scale). A few anomalous time series are highlighted as large dots.}
         \label{fig:teste_mu_EM}
\end{figure}

\section{Goodness of Fit}
\label{sec:goodness}

\begin{figure}
\centering
        \begin{subfigure}[b]{0.25\textwidth}
                 \includegraphics[width=\textwidth]{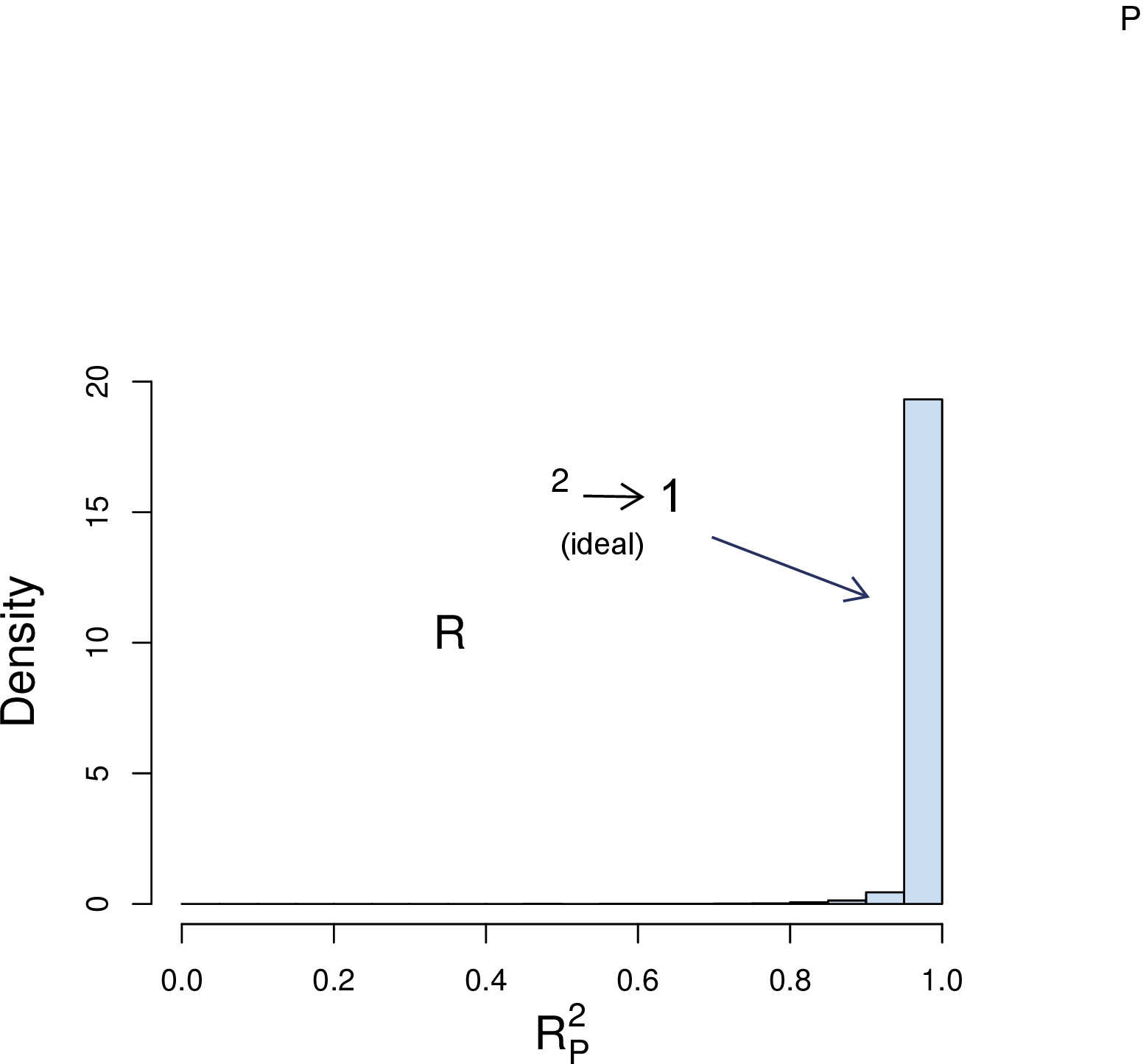}
                \caption{PP}
                \label{fig:Final_hist_r2pp}
       \end{subfigure}%
%        ~ %add desired spacing between images, e. g. ~, \quad, \qquad, \hfill etc.
%           %(or a blank line to force the subfigure onto a new line)
        \begin{subfigure}[b]{0.25\textwidth}
                 \includegraphics[width=\textwidth]{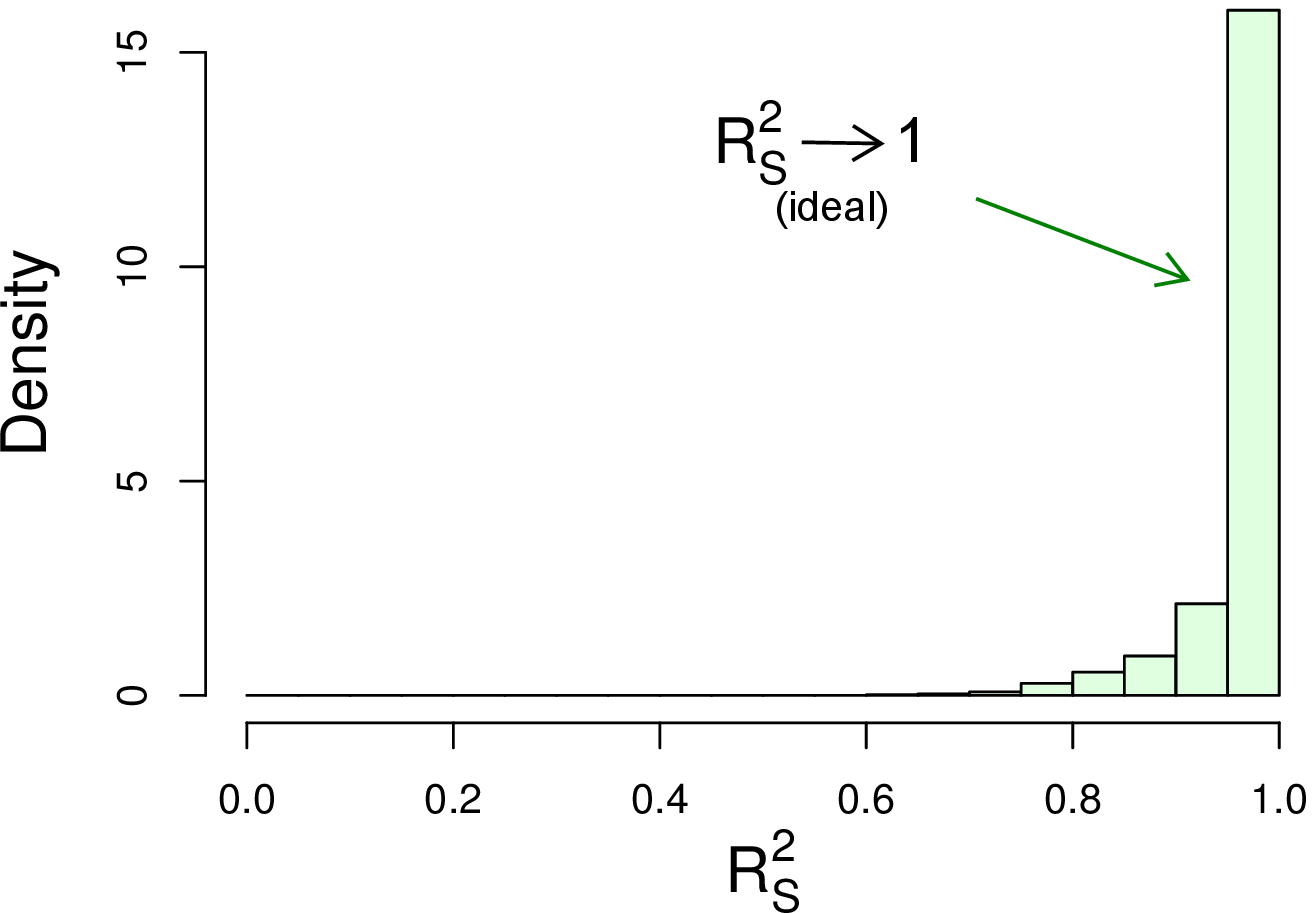}
                \caption{SFP}
                 \label{fig:Final_hist_r2SFP}
         \end{subfigure}
    \caption{Goodness of fit of mixed series}
    \label{fig:goodnessoffit}
%         ~ %add desired spacing between images, e. g. ~, \quad, \qquad, \hfill etc.
\end{figure}

Figure \ref{fig:goodnessoffit} shows a goodness of fit statistic for the RSE classified as 
a mixed process. After obtaining the $\hat{\lambda}_{p}$ and $\hat{\mu}$ estimates, we disentangled 
the two processes using the Monte Carlo simulation procedure described in Section \ref{subsec:muEMSec}.
The separated out events were then used to calculate the statistics shown in the two histograms.
The plot in Figure \ref{fig:Final_hist_r2pp} is the determination coefficient $R^2$ from the 
linear regression of the events cumulative number $N(t)$ versus $t$, which should be approximately
a straight line under the Poisson process process. 

In Figure \ref{fig:Final_hist_r2SFP} we show the $R^2$ from a linear regression with the 
SFP-labelled events. We take the inter-event times sample and build the empirical cumulative 
distribution function $\mathbb{F}(t)$ leading to the odds-ratio function 
$OR(t) = \mathbb{F}(t)/(1-\mathbb{F}(t))$. This function should be approximately a straight line
if the SFP process hypothesis is valid (more details in \cite{VazdeMelo2015}). 

Indeed, the two histograms of Figure \ref{fig:goodnessoffit} show very high concentration 
of the $R^2$ statistics close to the maximum value of 1 for the collection of RSE.
This provides evidence that our disentangling procedure of the mixed process into two components is 
able to create two processes that fir the characteristics of a Poisson process and a SFP process. 

\section{Comparison with Hawkes process}
\label{sec:Hawkes}

An alternative process to our model, is the Hawkes point process \cite{Crane2008},
which has conditional intensity 
defined by 
\begin{equation}
  \lambda(t | \mathcal{H}_t) = \lambda_p + \sum_{t_i < t} K(t - t_i)
    \label{eq:hawkess}
\end{equation} 
where $K(x) > 0$ is called the kernel function.
%, restricted to $\int_0^{\infty} K(x) dx < 1$ to guarantee that the process is stationary. 
As our \mixed model, the Hawkes process allows the successive events to interact 
with each other. However, there are two important differences between them. In Hawkes, every single event 
excites the process increasing the chance of additional events immediately after, while only some of these 
incoming events induces process excitement in BuSca. Depending on the value of $\paramScale$, only a fraction
of the events lead to an increase on $\lambda(t | \mathcal{H}_t)$. The second difference is the need to specify a 
functional form for the kernel $K(x)$, common choices being $K(x)$ with an exponential decay 
or a power law decay. 

We compared our model with the alternative Hawkes process using the 31431 events time series of all databases 
we analysed. We fitted both,\mixed and the Hawkes process, at each time series separately by maximum likelihood,
and evaluated the resulting Akaike information criterion. The Hawkes process was fitted with the exponential kernel
implying on a three parameter model while \mixed requires only two. The result was:  only four out of 31431 RSEs  
had their Hawkes AIC smaller. Therefore, for practically all time series we considered, 
our model fits better the data although requiring less parameters. 

%\begin{figure}[htp]
% \includegraphics[width=0.3\textwidth]{img/AIC.pdf}  
%          \caption{Akaike information criterion (AIC) of the two models, BuSca and Hawkes. 
%          Each dot represents one time series.}
%         \label{fig:AIC}
%\end{figure}

To understand better its relative failure,  we looked at the $R^2$ of the fitted Hawkes model
for each time series (see \cite{Peng2003} for the $R^2$ calculation in the Hawkess model). We studied the 
$R^2$ distribution conditioned on the value of $\paramScale$. Hawkes is able to fit reasonably well 
only when $\paramScale < 30\%$, that is, when the series of events is Poisson dominated.
It has mixed results for $0.3 \leq \paramScale \leq 0.5$, and a poor fit when  $\paramScale > 0.5$,
exactly when the bursty periods are more prevalent. 

\section{Applications}
\label{sec:aplications}

In this section, we described two applications based on our proposed model: the detection of series of events that should be 
considered as anomalous, given the typical statistical behavior of the population of series in a database; the detection of 
periods of bursts, when the series has a cascade of events that is due mainly to the SFP component.      

%%%%%%%%%%%%%%%%%%%%%%%%%%%%%%%%%%%%%%%%%%%%%%
\subsection{Anomaly detection}
 \label{subsec:anomalydetection}

Within a given a database, we saw empirically that the maximum likelihood estimator 
$(\log \hat{\lambda}_{p}$, $\log \hat{\mu})$ of the series of events follows approximately 
a bivariate Gaussian distribution. This can be justified by a Bayesian-type argument. Within a database, 
assume that the true parameters $(\lambda, \mu)$ (in the log scale) follow a bivariate Gaussian distribution. 
That is, each particular series has its own and specific parameter value $(\lambda, \mu)$. Conditional on this 
parameter vector, we know that the maximum likelihood estimator 
$(\log \hat{\lambda}_{p}, \log \hat{\mu})$ has an asymptotic distribution 
that is also a bivariate Gaussian. Therefore, unconditionally, the vector 
$(\log \hat{\lambda}_{p}, \log \hat{\mu})$ for the set of series of each database 
should exhibit a Gaussian behavior, as indeed we see in Figure \ref{fig:teste_mu_EM}.

Let $x_{ij} = (\log \hat{\lambda}_{pij}, \log \hat{\mu}_{ij})$ be the $i$-th individual of the $j$-th database.
We assume that the $x_{ij}$ from different individuals within a given database are i.i.d. bivariate random vectors 
following the bivariate Gaussian distribution $N_2(\mathbf{m}_j, \mathbf{\Sigma}_j)$ where $\mathbf{m}_j$ is the 
vector of expected values in database $j$ and $\mathbf{\Sigma}_j$ is its covariance matrix. 
To find the anomalous time series of events in database $j$, we used the Mahalanobis distance between 
each $j$-th individual and the typical value $\mathbf{m}_j$ given by:
\begin{equation}
D^2_{ij} = (\mathbf{x}_{ij} - \mathbf{m}_j)^{\prime} \mathbf{\Sigma}_j^{-1} (\mathbf{x}_{ij} - \mathbf{m}_j)
\label{eq:distance}
\end{equation}
Standard probability calculation establishes that $D^{2}_{ij}$ has a chi-square distribution when $x_{ij}$ is 
indeed selected from the bivariate Gaussian $N_2(\mathbf{m}_j, \mathbf{\Sigma}_j)$. This provides a direct 
score for an anomalous point time series. If its estimated vector $x_{ij}$ has the 
Mahalanobis distance $D^2_{ij} > c_{\alpha}$, the time series is considered anomalous. 
The threshold $c_{\alpha}$ is the $(1-\alpha)$-percentile of a chi-square distribution $\chi^2$, defined as 
$\mathbb{P}(\chi^2 > c_{\alpha}) = \alpha$. We adopted $\alpha = 0.01$. 

One additional issue in using (\ref{eq:distance}) is the unknown values for the expected vector $\mathbf{m}_j$
and the covariance matrix $\mathbf{\Sigma}$. We used robust estimates for these unknown parameters. Since we 
anticipate anomalous points among the sample, and we do not want them unduly affecting the estimates, we 
used estimation procedures that are robust to the presence of outliers. 
Specifically, we used the empirical medians in each database to estimate $\mathbf{m}_j$ and the 
median absolute deviation to estimate each marginal standard deviation. For the correlation parameter,
we substitute each robust marginal mean and standard deviation by its robust counterpart,
called correlation median estimator (see \cite{Smirnov2011}). 

\begin{figure}[H]
\centering
	      \includegraphics[width=0.50\textwidth]{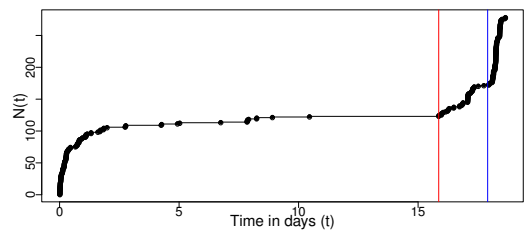}  
        \caption{Representation of topic $\#219940$, considered an anomaly in the \textit{AskMe} dataset.}
        \label{fig:219940}
\end{figure}

In Figure \ref{fig:teste_mu_EM} we highlighted 5 anomalous points found by our procedure to illustrate its 
usefulness. The first one correspond to the topic \#219940 of the \textit{AskMe} dataset, which has 
a very low value for $\hat{\lambda}_s$, compared to the other topics in this forum. This topic  
was initiated by a post about a lost dog and his owner asking for help. 
Figure \ref{fig:219940} shows the cumulative number $N(t)$ of events up to time $t$, measured in days. 
Consistent with the standard behavior in this forum, there is an initial burst of events 
with users suggesting ways to locate the pet or sympathizing with the pet owner. This is followed by 
a Poissonian period of events arising at a constant rate. Occasional bursts of lower intensity 
are still present but eventually the topic reaches a very low rate. Typically, about $t=12$ days, the topic 
would be considered dead and we would not see any additional activity. 
Further discussion from this point on would likely start a new topic. 
However, this was not what happened here. At $t=15.9$, the time marked by the vertical red line, 
the long inactivity period is broken by a post from the pet owner mentioning that he received 
new and promising information about the dog whereabouts. Once again, he receives a cascade of suggestions 
and supporting messages. Before this flow of events decreases substantially, he posts 
at time $t=17.9$ that the dog has been finally found. This is marked by the blue vertical line
and it caused a new cascade of 
events congratulating the owners by the good news. This topic is anomalous with respect to the others 
in the \textit{AskMe} database because, in fact, it contains three successive typical topics 
considered as a single one. 
The long inactivity period, in which the topic was practically dead, led to a $\hat{\lambda}_{p}$
with a very low value, reflecting mathematically the anomaly in the content we just described. 

In the \textit{MetaTalk} database, the time series \# 18067 and \# 21900 deal with a unusual topic in this 
platform. They are reminders of the deadline for posting in an semestral event among the users
and this prompted them to justify their lateness or make a comment about the event. 
This event is called \textit{MeFiSwap} and it a way found by the users to share their favorite playlists. 
The first one occurred in the summer of 2009 and the second one in the winter of 2012. 
Being reminders, they do not add content, but refer to and promote  other forum posts.
What is anomalous in these two time series is the time they took to develop: 22.6 days 
(\# 18067) and 10.9 (\# 21900), while the average topic takes about 1.9 days. The pattern 
within their enlarged time scale is the same as the rest of the database. 
The behavior of these two cases is closer to the Twitter population, as can be seen in  
Figure~\ref{fig:teste_mu_EM}.

The Twitter time series \# 1088 was pinpointed as an anomaly due to his relatively large value of the Poisson 
component $\lambda_p$. It offered free tickets for certain cultural event. To qualify for the tickets,
users should post something using the hashtag \texttt{iwantisatickets}. This triggered a cascade of associated 
posts that kept an approximately constant rate while it lasted.

Our final anomaly example is the time series \# 65232 from \textit{MetaFilter}. It was considered 
inappropriate and deleted from the forum by a moderator. The topic author suggested that grocery shopping 
should be exclusively a women's chore because his wife had discovered many deals he was unable to find out.
The subject was considered irrelevant and quickly prompted many criticisms among the users.

%%%%%%%%%%%%%%%%%%%%%%%%%%%%%%%%%%%%%%%%%%%%%%
\subsection{Burst detection and identification}
 \label{subsec:burstdetection}
 
Another practical application developed in this paper is the detection and identification of burst periods in each 
individual time series. This allows us to: (i) infer if a given topic is experiencing a quiet or a burst 
period; (ii) identify potential subtopics associated with distinct bursts; (iii) help understanding the causes 
of bursts.  The main idea is that a period with essentially no SFP activity should have  
the cumulative number of events $N(t)$ increasing at a constant and minimum slope approximately equal to 
$\lambda_p$. Periods with SFP activity would quickly increase this slope to some value $\lambda_p + c$.
We explore this intuitive idea by segmenting optimally the $N(t)$ series. 

\begin{figure*}
 \centering
        \begin{subfigure}[b]{0.30\textwidth}
  	  \includegraphics[width=\textwidth]{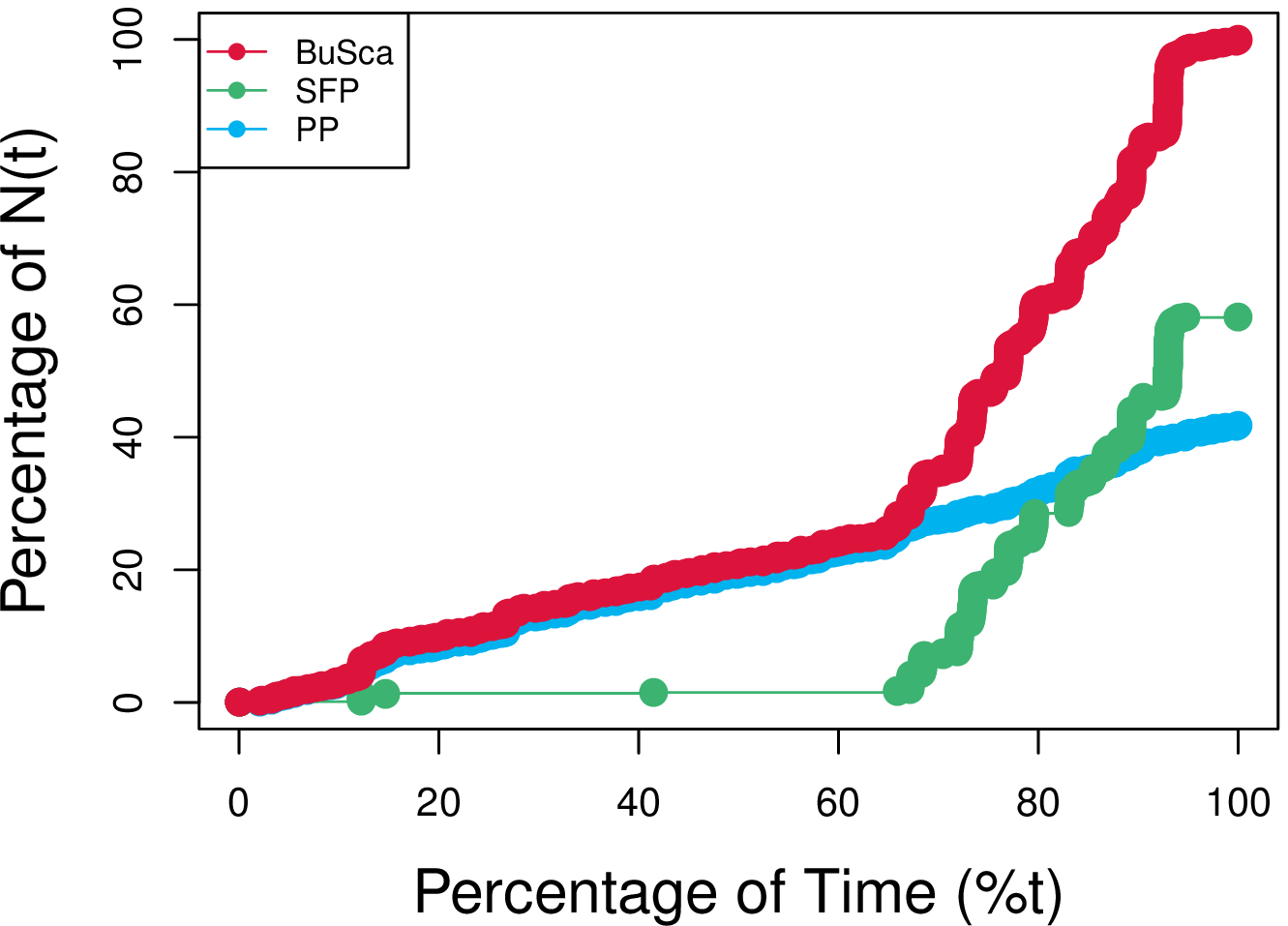}
                \caption{Cumulative series $N(t)$ versus $t$ (in percentage to their total).}
                \label{fig:10019_mix}
        \end{subfigure}%
        ~ %add desired spacing between images, e. g. ~, \quad, \qquad, \hfill etc.
        \begin{subfigure}[b]{0.30\textwidth}
               \includegraphics[width=\textwidth]{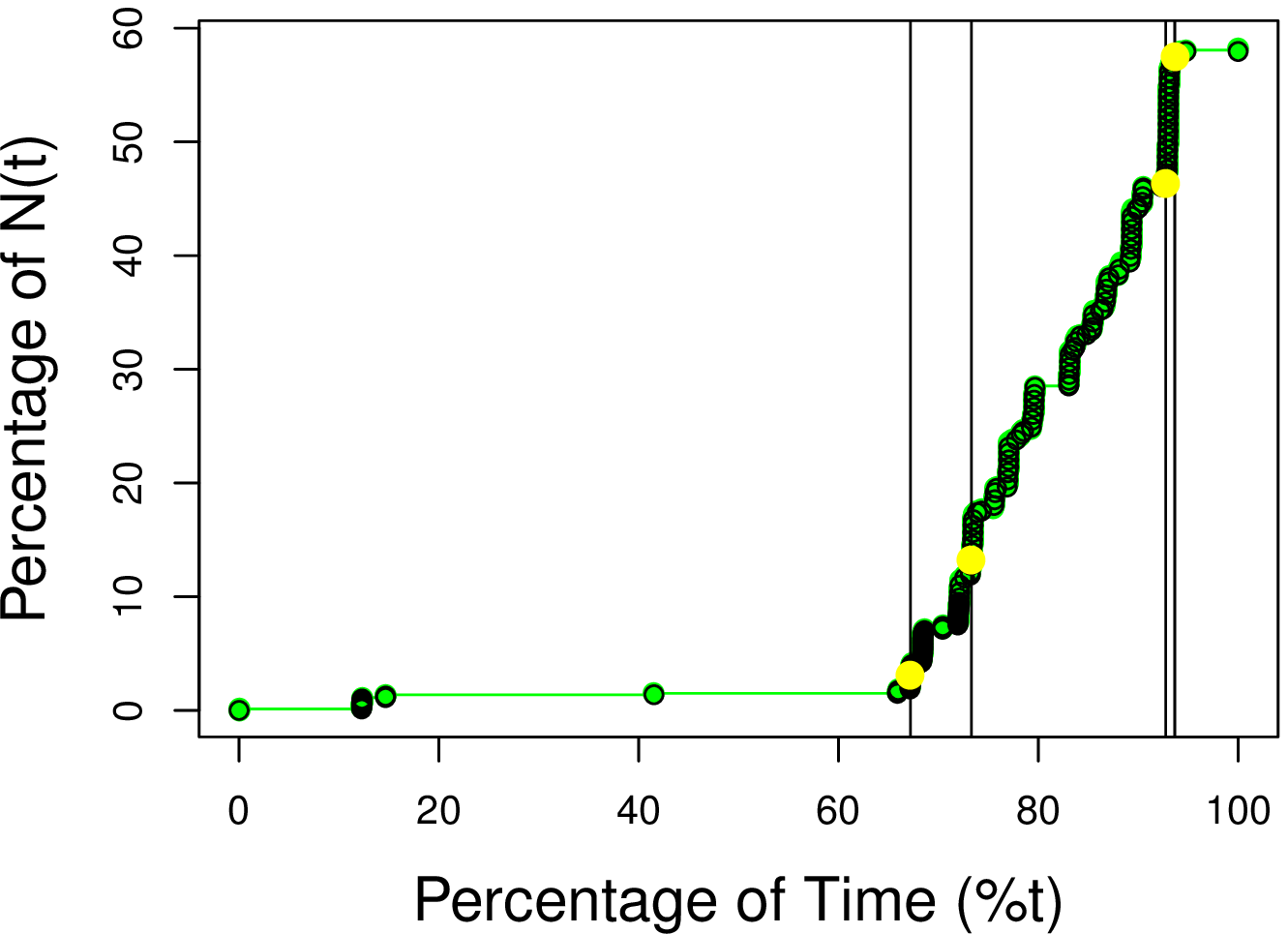}
                \caption{The SFP points used to carry out the apotimal segmentation.}
                \label{fig:10019_pontos}
        \end{subfigure}
        ~ %add desired spacing between images, e. g. ~, \quad, \qquad, \hfill etc.
        \begin{subfigure}[b]{0.30\textwidth}
                 \includegraphics[width=\textwidth]{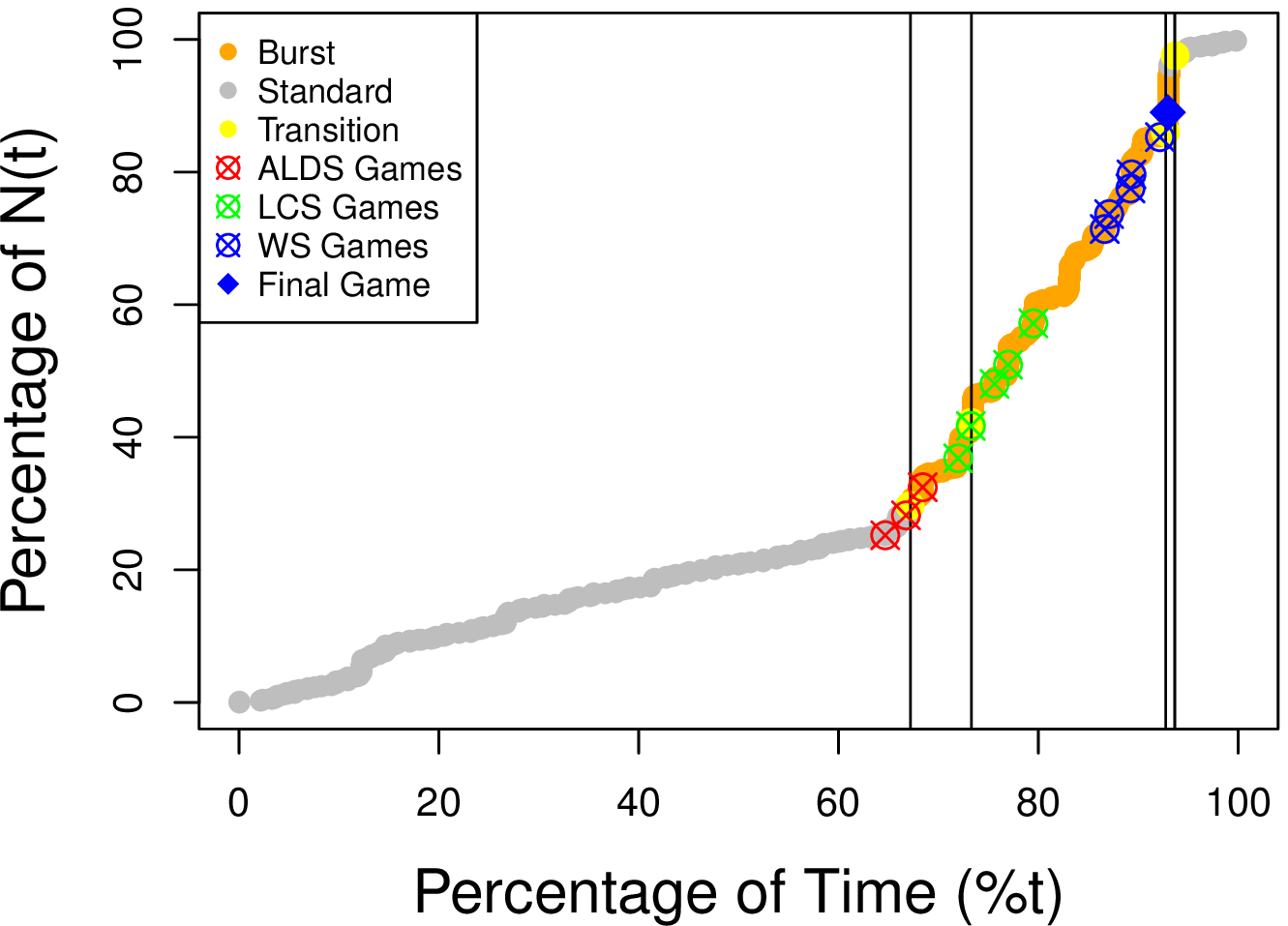}
                \caption{$N(t)$ with the optimal segments and the major games during playoffs.}
                \label{fig:10019_burst}
        \end{subfigure}
   \caption{Steps applying the burst detection algorithm for the individual hashtag \textit{\#Yankees!} in Twitter.}\label{fig:reddit_mix_ajuste}
 
\end{figure*}

We explain our method using Figure~\ref{fig:10019_mix}, showing the history of the Twitter hashtag
\texttt{\#Yankees!} (the red line) spanning the regular and postseason periods in 2009. 
We repeat several times the decomposition of the series of events into 
the two pure processes, Poisson and SFP, as explained in Section \ref{subsec:muEMSec}.
We select the best fitting one by considering a max-min statistics, the maximum over replications of the 
minimum $R^2$ of the two fits, the pure Poisson (blue line) and the pure SFP (green line). 
During the regular season, with posts coming essentially from the more enthusiastic fans, the behavior is 
completely dominated by a homogeneous Poisson process. 

Considering only the pure SFP events, we run the  
\textit{Segmented Least Squares} algorithm from \cite{kleinberg2006algorithm}.
For each potential segment, we fit a linear regression and obtain the
linear regression minimum sum of squares. A score measure for the segmentation is the sum 
over the segments of these minimum sum of squares. The best segmentation 
minimizes the score measure. Figure~\ref{fig:10019_pontos} shows the optimal segmentation 
of the \texttt{\#Yankees!} SFP series.   

This algorithm has $O(n^3)$ complexity on the number $n$ of points and therefore it is not efficient
for large time series. Hence, we reduced the number of 
SFP events by breaking them into 200 blocks or less. To avoid these 
blocks to be concentrated on burst periods, we mix two split strategies.
We selected 100 split points by taking the successive $k$-th percentiles (that is,
the event that leaves $100k\%$ of the events below it). We also divided the time segments into 100
equal length segments and took the closest event to each division point. These 200 points 
constitute the segments endpoints. 

The decision of which segment can be called a burst depends on the specific application. 
We say that each time segment $s = (t_i, t_f)$ found by the algorithm has a power $\tau(s)$, 
defined as the ratio between 
the observed number of SFP events and the expected number of Poisson events in the same segment. 
Therefore, the total number of points in a segment is approximately equal to $(\tau(s) + 1)\lambda_p$. 
The large the value of $\tau(s)$, the more intense the burst in that $s$ segment. For illustrative purposes,
we take $\tau(s) = 1$ as large enough to determine if the segment $s$ contains a SFP cascade.
In these cases, the segment has twice as much events as expected solely by the Poisson process. 

Figure~\ref{fig:10019_burst} returns to the original time series superimposing the segments division
and adding the main New York Yankees games during playoffs (the American League Division Series(ALDS), 
League Championship Series(LCS), and the World Series(WS)). The first segment found by the algorithm 
starts during the October 7 week, at the first postseason games against Minnesota Twins (red crosses in 
Figure~\ref{fig:10019_burst}). In this first playoff segment, we have $\tau(s) \approx 3$, or four times 
the standard regular behavior. The LCS games start an augmented burst until
November 4, when the New York Yankees defeated the Philadelphia Phillies in a final game.
This last game generated a very short burst marked by the blue diamond, with $\tau(s) \approx 64$.
After this explosive period, the series resume to the usual standard behavior. 

Analysing the $\tau(s)$ statistical distribution for each database separately, we found 
that they are well fitted by a heavy tailed probability distribution,  a finding that is 
consistent with previous studies of cascade events
(\cite{barabasi:2005}).

\section{Discussion and conclusions}
\label{sec:conclusions}
In this paper, we proposed the \mixedext (\mixed) model, which views each random series of events (RSEs) as a mix of two independent process: a \textit{Poissonian} and a self-exciting one. We presented and validated a particular and highly parsimonious case of \mixed, where the Poissonian process is given by a homogeneous Poisson process (PP) and the self-exciting process is given by a Self-Feeding Process (SFP)~\cite{vazdemelo:2013a}. When constructed in this way, \mixed is highly parsimonious, requiring only \textbf{two} parameters to characterize RSEs, one for the PP and another for the SFP. We validated our approach by analyzing eight diverse and large datasets containing real random series of events seen in Twitter, Yelp, e-mail conversations, Digg, and online forums. We also proposed a method that uses the \mixed model to disentangle events related to routine and constant behavior (Poissonian) from bursty and trendy ones (self-exciting). Moreover, from the two parameters of \mixed, we can calculate the burstiness scale parameter $\paramScale$, which represents how much of the RSE is due to bursty and viral effects. We showed that these two parameters, together with our proposed burstiness scale, is a highly parsimonious way to accurately characterize random series of events, which and, consequently, may leverage several applications, such as monitoring systems, anomaly detection methods, flow predictors, among others.

%\section{References}
% Either type in your references using
% \begin{thebibliography}{}
% Text
% \end{thebibliography}
%
% OR
%
% Compile your BiBTeX database using our plos2015.bst
% style file and paste the contents of your .bbl file
% here.
% 


\begin{thebibliography}{10}

\bibitem{Backstrom2013}
Lars Backstrom, Jon Kleinberg, Lillian Lee, and Cristian
  Danescu-Niculescu-Mizil.
\newblock {Characterizing and curating conversation threads}.
\newblock In {\em Proceedings of the sixth ACM international conference on Web
  search and data mining - WSDM '13}, page~13, New York, New York, USA, 2013.
  ACM Press.

\bibitem{barabasi:2005}
Albert-L{\'{a}}szl{\'{o}} Barab{\'{a}}si.
\newblock {The origin of bursts and heavy tails in human dynamics}.
\newblock {\em Nature}, 435(7039):207--211, may 2005.

\bibitem{ICWSM1510505}
Christian Bauckhage, Fabian Hadiji, and Kristian Kersting.
\newblock {How Viral Are Viral Videos?}, 2015.

\bibitem{Broxton2013}
Tom Broxton, Yannet Interian, Jon Vaver, and Mirjam Wattenhofer.
\newblock {Catching a viral video}.
\newblock {\em Journal of Intelligent Information Systems}, 40(2):241--259, apr
  2013.

\bibitem{Cao:2001}
Jin Cao, William~S. Cleveland, Dong Lin, and Don~X. Sun.
\newblock {Internet Traffic Tends Toward Poisson and Independent as the Load
  Increases}.
\newblock pages 83--109. 2003.

\bibitem{Choi2015}
Daejin Choi, Jinyoung Han, Taejoong Chung, Yong-Yeol Ahn, Byung-Gon Chun, and
  Ted~Taekyoung Kwon.
\newblock {Characterizing Conversation Patterns in Reddit}.
\newblock In {\em Proceedings of the 2015 ACM on Conference on Online Social
  Networks - COSN '15}, pages 233--243, New York, New York, USA, 2015. ACM
  Press.

\bibitem{Crane2008}
R.~Crane and D.~Sornette.
\newblock {Robust dynamic classes revealed by measuring the response function
  of a social system}.
\newblock {\em Proceedings of the National Academy of Sciences},
  105(41):15649--15653, oct 2008.

\bibitem{Du2015}
Nan Du, Mehrdad Farajtabar, Amr Ahmed, Alexander~J. Smola, and Le~Song.
\newblock {Dirichlet-Hawkes Processes with Applications to Clustering
  Continuous-Time Document Streams}.
\newblock In {\em Proceedings of the 21th ACM SIGKDD International Conference
  on Knowledge Discovery and Data Mining - KDD '15}, pages 219--228, New York,
  New York, USA, 2015. ACM Press.

\bibitem{eckmann:2004}
Jean-Pierre Eckmann, Elisha Moses, and Danilo Sergi.
\newblock {Entropy of dialogues creates coherent structures in e-mail traffic}.
\newblock {\em Proceedings of the National Academy of Sciences of the United
  States of America}, 101(40):14333--14337, 2004.

\bibitem{FerrazCosta2015}
Alceu {Ferraz Costa}, Yuto Yamaguchi, Agma {Juci Machado Traina}, Caetano
  Traina, and Christos Faloutsos.
\newblock {RSC}.
\newblock In {\em Proceedings of the 21th ACM SIGKDD International Conference
  on Knowledge Discovery and Data Mining - KDD '15}, pages 269--278, New York,
  New York, USA, 2015. ACM Press.

\bibitem{Figueiredo2014}
Flavio Figueiredo, Jussara~M. Almeida, Yasuko Matsubara, Bruno Ribeiro, and
  Christos Faloutsos.
\newblock {Revisit Behavior in Social Media: The Phoenix-R Model and
  Discoveries}.
\newblock pages 386--401. 2014.

\bibitem{Filimonov2015}
Vladimir Filimonov, Spencer Wheatley, and Didier Sornette.
\newblock {Effective measure of endogeneity for the Autoregressive Conditional
  Duration point processes via mapping to the self-excited Hawkes process}.
\newblock {\em Communications in Nonlinear Science and Numerical Simulation},
  22(1-3):23--37, may 2015.

\bibitem{Gao2015}
Shuai Gao, Jun Ma, and Zhumin Chen.
\newblock {Modeling and Predicting Retweeting Dynamics on Microblogging
  Platforms}.
\newblock In {\em Proceedings of the Eighth ACM International Conference on Web
  Search and Data Mining - WSDM '15}, pages 107--116, New York, New York, USA,
  feb 2015. ACM Press.

\bibitem{Garriss:2006}
Scott Garriss, Michael Kaminsky, Michael~J Freedman, Brad Karp, David
  Mazi{\`{e}}res, and Haifeng Yu.
\newblock {Re: Reliable Email}.
\newblock In {\em Proceedings of the Third USENIX/ACM Symposium on Networked
  System Design and Implementation (NSDI'06)}, pages 297--310, 2006.

\bibitem{Gomez2013}
Vicen{\c{c}} G{\'{o}}mez, Hilbert~J. Kappen, Nelly Litvak, and Andreas
  Kaltenbrunner.
\newblock {A likelihood-based framework for the analysis of discussion
  threads}.
\newblock {\em World Wide Web}, 16(5-6):645--675, nov 2013.

\bibitem{haight:1967}
Frank~A Haight.
\newblock {\em {Handbook of the Poisson distribution [by] Frank A. Haight}}.
\newblock Wiley New York,, 1967.

\bibitem{Jiang:2005}
Hao Jiang and Constantinos Dovrolis.
\newblock {Why is the Internet Traffic Bursty in Short Time Scales?}
\newblock In {\em Proceedings of the 2005 ACM SIGMETRICS International
  Conference on Measurement and Modeling of Computer Systems (SIGMETRICS'05)},
  pages 241--252, 2005.

\bibitem{Karagiannis:2004}
T.~Karagiannis, M.~Molle, M.~Faloutsos, and A.~Broido.
\newblock {A nonstationary poisson view of internet traffic}.
\newblock In {\em IEEE INFOCOM 2004}, volume~3, pages 1558--1569. IEEE.

\bibitem{kleinberg:2002}
Jon Kleinberg.
\newblock {Bursty and hierarchical structure in streams}.
\newblock In {\em Proceedings of the eighth ACM SIGKDD}, KDD '02, pages
  91--101, New York, NY, USA, 2002. ACM.

\bibitem{kleinberg2006algorithm}
Jon Kleinberg and Eva Tardos.
\newblock {\em Algorithm design}.
\newblock Pearson Education, 2006.

\bibitem{Lehmann2012}
Janette Lehmann, Bruno Gon{\c{c}}alves, Jos{\'{e}}~J. Ramasco, and Ciro
  Cattuto.
\newblock {Dynamical classes of collective attention in twitter}.
\newblock In {\em Proceedings of the 21st international conference on World
  Wide Web - WWW '12}, page 251, New York, New York, USA, 2012. ACM Press.

\bibitem{Lerman2010}
Kristina Lerman and Tad Hogg.
\newblock {Using a model of social dynamics to predict popularity of news}.
\newblock In {\em Proceedings of the 19th international conference on World
  wide web - WWW '10}, page 621, New York, New York, USA, 2010. ACM Press.

\bibitem{malmgren:2008}
R.~D. Malmgren, D.~B. Stouffer, A.~E. Motter, and L.~A.~N. Amaral.
\newblock {A Poissonian explanation for heavy tails in e-mail communication}.
\newblock {\em Proceedings of the National Academy of Sciences},
  105(47):18153--18158, nov 2008.

\bibitem{malmgren:2009}
R.~Dean Malmgren, Jake~M. Hofman, Luis~A.N. Amaral, and Duncan~J. Watts.
\newblock {Characterizing individual communication patterns}.
\newblock In {\em Proceedings of the 15th ACM SIGKDD international conference
  on Knowledge discovery and data mining - KDD '09}, page 607, New York, New
  York, USA, 2009. ACM Press.

\bibitem{Masuda2013}
Naoki Masuda, Taro Takaguchi, Nobuo Sato, and Kazuo Yano.
\newblock {Self-Exciting Point Process Modeling of Conversation Event
  Sequences}.
\newblock In {\em Understanding Complex Systems}, chapter Temporal N, pages
  245--264. 2013.

\bibitem{Matsubara2012}
Yasuko Matsubara, Yasushi Sakurai, B.~Aditya Prakash, Lei Li, and Christos
  Faloutsos.
\newblock {Rise and fall patterns of information diffusion}.
\newblock In {\em Proceedings of the 18th ACM SIGKDD international conference
  on Knowledge discovery and data mining - KDD '12}, page~6, New York, New
  York, USA, 2012. ACM Press.

\bibitem{oliveira:2005}
Joao~G Oliveira and Albert-Laszlo Barabasi.
\newblock {Human dynamics: Darwin and Einstein correspondence patterns}.
\newblock {\em Nature}, 437(7063):1251, 2005.

\bibitem{Peng2003}
Roger Peng.
\newblock Multi-dimensional point process models in r.
\newblock {\em Journal of Statistical Software}, 8(1):1--27, 2003.

\bibitem{Pinto2015}
Julio Cesar~Louzada Pinto, Tijani Chahed, and Eitan Altman.
\newblock {Trend detection in social networks using Hawkes processes}.
\newblock In {\em Proceedings of the 2015 IEEE/ACM International Conference on
  Advances in Social Networks Analysis and Mining 2015 - ASONAM '15}, pages
  1441--1448, New York, New York, USA, 2015. ACM Press.

\bibitem{Romero2011}
Daniel~M. Romero, Brendan Meeder, and Jon Kleinberg.
\newblock {Differences in the mechanics of information diffusion across
  topics}.
\newblock In {\em Proceedings of the 20th international conference on World
  wide web - WWW '11}, page 695, New York, New York, USA, 2011. ACM Press.

\bibitem{Smirnov2011}
G.~Shevlyakov and P.~Smirnov.
\newblock {Robust Estimation of the Correlation Coefficient: An Attempt of
  Survey}.
\newblock {\em Austrian Journal of Statistics}, 40(1):147--156, 2011.

\bibitem{Siersdorfer2014}
Stefan Siersdorfer, Sergiu Chelaru, Jose~San Pedro, Ismail~Sengor Altingovde,
  and Wolfgang Nejdl.
\newblock {Analyzing and Mining Comments and Comment Ratings on the Social
  Web}.
\newblock {\em ACM Transactions on the Web}, 8(3):1--39, jul 2014.

\bibitem{Snyder1991}
Donald~L. Snyder and Michael~I. Miller.
\newblock {\em {Random Point Processes in Time and Space}}.
\newblock Springer Texts in Electrical Engineering. Springer New York, New
  York, NY, 1991.

\bibitem{Spasojevic2015}
Nemanja Spasojevic, Zhisheng Li, Adithya Rao, and Prantik Bhattacharyya.
\newblock {When-To-Post on Social Networks}.
\newblock In {\em Proceedings of the 21th ACM SIGKDD International Conference
  on Knowledge Discovery and Data Mining - KDD '15}, pages 2127--2136, New
  York, New York, USA, 2015. ACM Press.

\bibitem{Vallet:2015:CPV:2806416.2806556}
David Vallet, Shlomo Berkovsky, Sebastien Ardon, Anirban Mahanti, and
  Mohamed~Ali Kafaar.
\newblock {Characterizing and Predicting Viral-and-Popular Video Content}.
\newblock In {\em Proceedings of the 24th ACM International on Conference on
  Information and Knowledge Management}, CIKM '15, pages 1591--1600, New York,
  NY, USA, 2015. ACM.

\bibitem{vazdemelo:2013a}
Pedro O~S {Vaz de Melo}, Christos Faloutsos, Renato Assuncao, and Antonio A~F
  Loureiro.
\newblock {The Self-Feeding Process: A Unifying Model for Communication
  Dynamics in the Web}.
\newblock In {\em WWW '13: 22nd International World Wide Web Conference}, 2013.

\bibitem{VazdeMelo2015}
Pedro Olmo~Stancioli {Vaz de Melo}, Christos Faloutsos, Renato
  Assun{\c{c}}{\~{a}}o, Rodrigo Alves, and Antonio~A.F. Loureiro.
\newblock {Universal and Distinct Properties of Communication Dynamics: How to
  Generate Realistic Inter-event Times}.
\newblock {\em ACM Transactions on Knowledge Discovery in Data}, 2015.

\bibitem{vazques:2006}
Alexei Vazquez, Joao~Gama Oliveira, Zoltan Dezso, Kwang-Il Goh, Imre Kondor,
  and Albert-Lazlo Barabasi.
\newblock {Modeling bursts and heavy tails in human dynamics}.
\newblock {\em Phys Rev E Stat Nonlin Soft Matter Phys}, 73:36127, 2006.

\bibitem{Wang2012}
Chunyan Wang, Mao Ye, and Bernardo~A. Huberman.
\newblock {From user comments to on-line conversations}.
\newblock In {\em Proceedings of the 18th ACM SIGKDD international conference
  on Knowledge discovery and data mining - KDD '12}, page 244, New York, New
  York, USA, 2012. ACM Press.

\bibitem{AAAI159338}
Senzhang Wang, Zhao Yan, Xia Hu, Philip~S Yu, and Zhoujun Li.
\newblock {Burst Time Prediction in Cascades}, 2015.

\bibitem{Yang2011}
Jaewon Yang and Jure Leskovec.
\newblock {Patterns of temporal variation in online media}.
\newblock In {\em Proceedings of the fourth ACM international conference on Web
  search and data mining - WSDM '11}, page 177, New York, New York, USA, 2011.
  ACM Press.

\bibitem{yang13a}
Shuang-hong Yang and Hongyuan Zha.
\newblock {Mixture of Mutually Exciting Processes for Viral Diffusion}.
\newblock In Sanjoy Dasgupta and David Mcallester, editors, {\em Proceedings of
  the 30th International Conference on Machine Learning (ICML-13)}, volume~28,
  pages 1--9. JMLR Workshop and Conference Proceedings, 2013.

\bibitem{ICWSM1510537}
Honglin Yu, Lexing Xie, and Scott Sanner.
\newblock {The Lifecyle of a Youtube Video: Phases, Content and Popularity},
  2015.

\bibitem{Zhao2015}
Qingyuan Zhao, Murat~A. Erdogdu, Hera~Y. He, Anand Rajaraman, and Jure
  Leskovec.
\newblock {SEISMIC: A Self-Exciting Point Process Model for Predicting Tweet
  Popularity}.
\newblock In {\em Proceedings of the 21th ACM SIGKDD International Conference
  on Knowledge Discovery and Data Mining - KDD '15}, pages 1513--1522, New
  York, New York, USA, 2015. ACM Press.

\end{thebibliography}
\end{document}